\documentclass[10pt]{article} %
\usepackage[preprint]{tmlr}

\usepackage{hyperref}
\colorlet{citeblue}{blue!50!black}
\colorlet{linkred}{red!50!black}
\hypersetup{
  colorlinks,
  linkcolor=citeblue, %
  citecolor=citeblue,
  urlcolor=citeblue,
  breaklinks=true,   %
  linktoc=all,
}

\usepackage{url}

\usepackage{amsmath}
\usepackage{amsthm}
\usepackage{amssymb}

\usepackage{algorithm}
\usepackage{algpseudocode}

\usepackage{graphicx}
\usepackage{booktabs}
\usepackage{makecell}

\usepackage{pgf}
\usepackage{pgfplots}
\pgfplotsset{compat=newest}

\newcommand{\expm}[1]{{#1}} %

\newcommand{\defm}[1]{\emph{#1}} %

\newcommand{\setDec}{D} %
\newcommand{\safeLevel}{h} %
\newcommand{\model}{\mathcal{M}} %
\newcommand{\setExp}{G} %
\newcommand{\setMax}{M} %

\newcommand{\K}{\mathbf{K}} %
\newcommand{\Ident}{\mathbf{I}} %
\newcommand{\Hpre}{H^{\text{pre}}} %

\newcommand{\R}{\mathbb{R}} %
\newcommand{\Rp}{\mathbb{R}_{>0}} %
\newcommand{\Rnn}{\mathbb{R}_{\geq 0}} %
\newcommand{\N}{\mathbb{N}_0} %
\newcommand{\Np}{\mathbb{N}_{>0}} %
\newcommand{\argmax}{\mathrm{argmax}}

\newcommand{\vspan}{\mathrm{span}} %

\newcommand{\GP}{\mathcal{GP}}
\newcommand{\E}{\mathbb{E}} %
\newcommand{\Pb}{\mathbb{P}} %

\newtheorem{proposition}{Proposition}
\newtheorem{remark}{Remark}
\newtheorem{assumption}{Assumption}

\title{On Safety in Safe Bayesian Optimization}

\author{%
\name Christian Fiedler$^\ast$ \email fiedler@dsme.rwth-aachen.de \\
\addr Institute for Data Science in Mechanical Engineering (DSME) \\ RWTH Aachen University
\AND 
\name Johanna Menn$^\ast$ \email johanna.menn@dsme.rwth-aachen.de \\
\addr Institute for Data Science in Mechanical Engineering (DSME) \\ RWTH Aachen University
\AND 
\name Lukas Kreisköther
\AND 
\name Sebastian Trimpe \email trimpe@dsme.rwth-aachen.de \\
\addr Institute for Data Science in Mechanical Engineering (DSME) \\ RWTH Aachen University
\AND
{\normalfont{(* Equal contribution)}} 
}

\def\month{MM}  %
\def\year{YYYY} %
\def\openreview{\url{https://openreview.net/forum?id=XXXX}} %

\begin{document}
\maketitle
\begin{abstract}
Optimizing an unknown function under safety constraints is a central task in robotics, biomedical engineering, and many other disciplines,
and increasingly safe Bayesian Optimization (BO) is used for this.
Due to the safety critical nature of these applications, it is of utmost importance that theoretical safety guarantees for these algorithms translate into the real world.
In this work, we investigate three safety-related issues of the popular class of SafeOpt-type algorithms.
First, these algorithms critically rely on frequentist uncertainty bounds for Gaussian Process (GP) regression,
but concrete implementations typically utilize heuristics that invalidate all safety guarantees. 
We provide a detailed analysis of this problem and introduce Real-$\beta$-SafeOpt, a variant of the SafeOpt algorithm that leverages recent GP bounds and thus retains all theoretical guarantees. 
Second, we identify assuming an upper bound on the reproducing kernel Hilbert space (RKHS) norm of the target function, a key technical assumption in SafeOpt-like algorithms, as a central obstacle to real-world usage.
To overcome this challenge, we introduce the Lipschitz-only Safe Bayesian Optimization (LoSBO) algorithm, 
which guarantees safety without an assumption on the RKHS bound,
and empirically show that this algorithm is not only safe, but also exhibits superior performance compared to the state-of-the-art on several function classes. 
Third, SafeOpt and derived algorithms rely on a %
discrete search space, making them difficult to apply to higher-dimensional problems.
To widen the applicability of these algorithms, we introduce Lipschitz-only GP-UCB (LoS-GP-UCB), a variant of LoSBO applicable to moderately high-dimensional problems, while retaining safety.
By analysing practical safety issues of an important class of safe BO algorithms, and providing ready-to-use algorithms overcoming these problems, the present work helps to bring safe and reliable machine learning techniques closer to real world applications.
\end{abstract}

\section{Introduction}
A frequent task in science, engineering, and business is optimizing an unknown target function that is expensive to evaluate, and for we have only noisy function values available. If it is possible to actively query the function, i.e., to select the inputs that are evaluated, this problem is commonly addressed using Bayesian Optimization (BO), see \citet{Shahriari2015} or \citet{garnett2023bayesian} for an overview. However, in many real-world applications, BO algorithms should avoid using certain inputs, often due to safety considerations. 
For example, the target function might be a reward function for a robotic task and the input is a control policy for a physical robot. 
Inputs that lead to physical damage or unsafe behavior should then be avoided. An important special case of such a safety constraint is the requirement to allow only query inputs with function values not lower than a given threshold \citep{kim2020safe}. In such a case, a safe input is one that leads to a function value above a given threshold, and a BO algorithm is called safe if it does not query unsafe inputs throughout its run. This type of safety constraint has been introduced in \citet{sui2015safe} and arises for example in biomedical applications (where the target function is patient comfort and the input corresponds to treatment settings) or controller tuning (where the target function is a measure of controller performance and the inputs are tuning parameters).

A popular BO algorithm for this problem setting is SafeOpt \citep{sui2015safe}. Starting from a given set of safe inputs, this algorithm iteratively searches for a maximum while %
aiming to avoid unsafe inputs with high probability
It achieves this by utilizing Gaussian Processes (GPs) together with a frequentist uncertainty bound \citep{Srinivas2010,Chowdhury2017} \emph{and} a known upper bound on the Lipschitz constant of the target function. 
Provided the algorithmic parameters are set correctly, then with (high) pre-defined probability, SafeOpt demonstrably converges to the safely reachable maximum while avoiding unsafe inputs. 
SafeOpt and its variants have been used in various applications, e.g., safe controller optimization \citep{berkenkamp2016safe}, automated deep brain stimulation \citep{Sarikhani2021} and safe robot learning \citep{baumann2021gosafe}.

To ensure safety, SafeOpt and its variants require frequentist uncertainty sets that are valid (i.e., holding with specified high probability) and explicit (i.e., they can be numerically evaluated). 
However, two issues arise here: First, the uncertainty bounds from \citet{Srinivas2010,Chowdhury2017} used in most SafeOpt-type algorithms tend to be conservative, even if all necessary ingredients for their computation are available. This can completely prevent exploration of the target function. 
Second, these uncertainty bounds rely upon a particular property of the target function (a known finite upper bound on the Reproducing Kernel Hilbert Space (RKHS) norm, cf. Section \ref{sec:rkhs} for details), which in practice is very difficult to derive from reasonable prior knowledge. As a consequence of these issues, algorithmically usable uncertainty sets have not yet been available for SafeOpt. In fact, to the best of our knowledge, all implementations of SafeOpt and its variants have used heuristics instead \citep{berkenkamp2016safe,kirschner2019adaptive,baumann2021gosafe,Koller2019,Helwa2019}.\footnote{The precise experimental settings are not reported in \citet{sui2015safe}, however, based on the descriptions in this work, it can be inferred that some form of heuristic was used.} This shortcoming means that such implementations lose all their theoretical safety guarantees in practice.
In this work, we carefully investigate this issue, and propose practical solutions, bringing safe BO closer to real-world usage.

\paragraph{Outline}
In Section \ref{sec:background}, necessary technical background on Gaussian Process (GP) regression, reproducing kernel Hilbert spaces (RKHSs), and frequentist uncertainty bounds for GP regression will be reviewed.
The problem setting and our objectives are presented in Section \ref{sec:problemSetting}, including a detailed description of the original SafeOpt algorithm.
We provide a comprehensive discussion of related work in Section \ref{sec:relatedWork}. %
We start our investigation into safety aspects of SafeOpt in Section \ref{sec:realBeta}, where we discuss practical problems arising from the use of heuristics in SafeOpt-type algorithms and demonstrate these issues with numerical experiments.
To overcome these problems, we propose to use state-of-the-art uncertainty bounds in the actual algorithms, leading to an algorithmic variant we call Real-$\beta$-SafeOpt, which also forms the foundation of the following numerical experiments.
In Section \ref{sec:losbo}, we discuss the central assumption of a known upper bound on the RKHS norm of the target function, and highlight the practical safety problems that arise from it.
This motivates the introduction of Lipschitz-only Safe Bayesian Optimization (LoSBO), a SafeOpt-type algorithm that does not rely on this assumption for safety.
Furthermore, we prove appropriate safety guarantees for LoSBO, and thoroughly evaluate the algorithm using numerical experiments.
To allow this type of safe BO also in high dimensions, in Section \ref{sec:highdim} another algorithmic variant called Lipschitz-only Safe Gaussian Process-Upper Confidence Bound (LoS-GP-UCB) is introduced.
We describe and discuss the algorithm in detail, and perform numerical experiments to evaluate it empirically.
Finally, Section \ref{sec:conclusion} closes the article with a summary and outlook.

\section{Background} \label{sec:background}
In this section, we provide a brief review of Gaussian Process (GP) regression, reproducing kernel Hilbert spaces (RKHSs), and frequentist uncertainty bounds for GP regression, since these three components form the foundations for SafeOpt-type algorithms.
In our review of frequentist uncertainty bounds for GP regression, we also point out a computable form of such an uncertainty bound, which is based on existing theory, but which seems to have been underutilized in the literature.
\subsection{Gaussian Processes} \label{sec:gps}
A GP is a collection of $\mathbb{R}$-valued random variables, here indexed by the set $\setDec$, such that every finite collection of those random variables has a multivariate normal distribution. 
A GP $g$ is uniquely defined by its mean function \(m(x)=\E[g(x)]\) and covariance function \(k(x,x')=\E[(g(x)-m(x))((g(x')-m(x'))]\), and we denote such a GP by \(g \sim \mathcal{GP}_\setDec(m,k)\). 
In GP regression, we start with a prior GP. Assuming independent and identically distributed (i.i.d.) additive normal noise, \(\epsilon_t \sim \mathcal{N}(0, \sigma^2)\), this prior GP can be updated with data $(x_1,y_1),\ldots,(x_t,y_t)$, leading to a posterior GP. Without loss of generality we assume that the prior GP has zero mean. Then the posterior mean, posterior covariance and posterior variance are given by
$\mu_t(x) = \boldsymbol{k}_t(x)^T (\boldsymbol{K}_t + \sigma^2 \boldsymbol{I}_t)^{-1} \boldsymbol{y}_t$,
$k_t(x,x') = k(x,x') - \boldsymbol{k}_t(x)^T (\boldsymbol{K}_t + \sigma^2 \boldsymbol{I}_t)^{-1} \boldsymbol{k}_t(x')$, and
$\sigma^2_t(x) = k_t(x,x)$, respectively. 
We also defined \(\boldsymbol{y}_t=[y_1,...,y_t]^T\) is the vector of observed, noisy function values of \(f\), the kernel matrix \(\boldsymbol{K}_t \in \mathbb{R}^{t \times t}\) has entries \([k(x,x')]_{x,x' \in D_t}\), the vector \(\boldsymbol{k}_t(x)=[k(x_1,x) \cdots k(x_t,x)]^T\) contains the covariances between \(x\) and the observed data points, and \(I_t\) is the \(t \times t\) identity matrix. 

In practice, the prior mean, prior covariance function, and noise level, are (partially) chosen based on prior knowledge.
For example, if no specific prior knowledge regarding the mean is available, usually the zero function is chosen.
Furthermore, in practice these three components are only partially specified, usually up to some parameters, which are then called hyperparameters in the context.
In BO, these are often determined during the optimization via hyperparameter optimization \citep{garnett2023bayesian}.
We will come back to this issue in Section \ref{sec:uncertaintyBounds} and carefully outline and justify our approach.

For more details on GPs, GP regression, and related method, we refer to \citet{Rasmussen2006} and \citet{garnett2023bayesian}.

\subsection{Reproducing kernel Hilbert spaces} \label{sec:rkhs}
Consider a function $k: \setDec\times\setDec\rightarrow\R$.
We call $k$ \defm{positive semidefinite} if for all $x_1,\ldots,x_N\in\setDec$, $N\in\Np$, the matrix $\begin{pmatrix} k(x_i,x_j) \end{pmatrix}_{i,j=1,\ldots,N}$ is positive semidefinite. Equivalently, the function $k$ is symmetric ($k(x_i,x_j)=k(x_j,x_i)$ for all $i,j=1,\ldots,N$), and for all $\alpha_1,\ldots,\alpha_N\in\R$ we have $\sum_{i,j=1}^N \alpha_i\alpha_j k(x_i,x_j)\geq 0$.
In the literature such a function is often called \defm{positive definite} or \defm{of positive type}.
Additionally, $k$ is called \defm{positive definite}, if for all pairwise distinct $x_1,\ldots,x_N\in\setDec$, the matrix $\begin{pmatrix} k(x_i,x_j) \end{pmatrix}_{i,j=1,\ldots,N}$ is positive definite. This property is sometimes called \defm{strict positive definiteness} in the literature.

Let $H$ be a Hilbert space functions on $\setDec$. We call $H$ a \defm{reproducing kernel Hilbert space (RKHS)} if every evaluation functional is continuous, i.e., for all $x\in\setDec$ the mapping $H\ni f \mapsto f(x)\in \R$ is continuous w.r.t. the norm induced by the scalar product in $H$.
Furthermore, $k$ is called a \defm{reproducing kernel (for $H$)} if
1) $k(\cdot,x)\in H$ for all $x\in\setDec$, and 2) $f(x)=\langle f, k(\cdot,x)\rangle_H$ for all $f\in H$ and $x\in \setDec$.

As is well-known, $H$ is a RKHS if and only if it has a reproducing kernel, and in this case the latter is unique \citep[Lemma~4.19,~Theorem~4.20]{SC08}.
Furthermore, every reproducing kernel is positive semidefinite, and every positive semidefinite function is a reproducing kernel for a unique RKHS \cite[Theorem~4.16,~4.21]{SC08}.
If $k$ is positive semidefinite, then we denote its unique RKHS as $(H_k,\langle\cdot,\cdot\rangle_k)$, and the induced norm by $\|\cdot\|_k$.
Furthermore, we define the \defm{pre RKHS} by
\begin{equation}
    \Hpre_k = \vspan\left\{ k(\cdot,x) \mid x\in\setDec \right\} = \left\{ \sum_{n=1}^N \alpha_n k(\cdot,x_n) \mid N\in\Np, \alpha_n\in\R, x_n\in\setDec,\: n=1,\ldots,N \right\},
\end{equation}
and this subspace is dense in $H_k$ w.r.t. $\|\cdot\|_k$. Given $f=\sum_{n=1}^N \alpha_n k(\cdot,x_n)$, $g=\sum_{m=1}^M \beta_m k(\cdot,y_m) \in \Hpre_k$, we have
\begin{equation}
    \langle f, g \rangle_k = \sum_{n=1}^N \sum_{m=1}^M \alpha_n \beta_m k(y_m,x_n),
\end{equation}
cf. \citet[Theorem~4.21]{SC08}.

If $k$ is the covariance function of a GP, then it is positive semidefinite (since every covariance matrix is positive semidefinite), and hence the reproducing kernel of a unique RKHS. Conversely, if $k$ is the reproducing kernel of a RKHS, then $k$ is positive semidefinite, and there exists a GP having $k$ as its covariance function, and the GP can be chosen with a zero mean function \citep{BTA04}.

Furthermore, consider GP regression with a prior $\GP_\setDec(m,k)$, then $\mu_t-m\in \Hpre_k$, where $\mu_t$ is the posterior mean corresponding to a finite data set with $t$ points.
In particular, the posterior mean for a zero mean prior GP is always in the pre RKHS corresponding to the covariance function.
As is customary in machine learning with GPs \citep{Rasmussen2006}, and also in many BO scenarios \citep{garnett2023bayesian}, in the following we will use without loss of generality a zero mean GP prior, $m\equiv 0$.
\subsection{Frequentist Uncertainty Bounds} \label{sec:uncertaintyBounds}
\begin{figure}
    \hfill
    \includegraphics[width=0.49\textwidth]{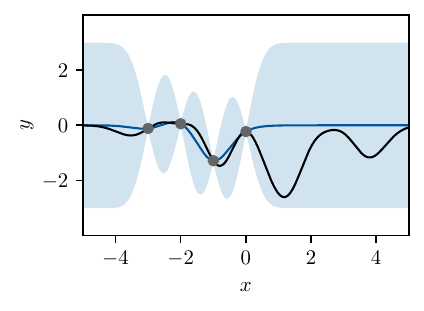}
    \hfill
    \includegraphics[width=0.49\textwidth]{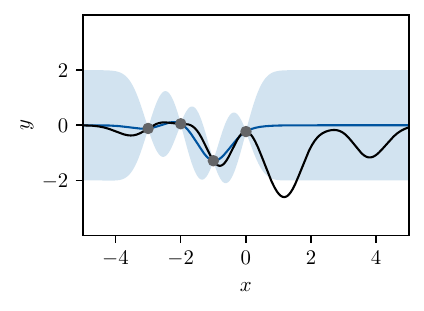}
    \hfill
    \caption{Illustration of the required GP error bounds. Consider a fixed ground truth (solid black line), of which only finitely many samples are known (blue crosses). Applying GP regression leads to a posterior GP, from which a high-probability uncertainty set can be derived (shaded blue).
    Left: The ground truth is completely contained in the uncertainty set. 
    Right: The ground truth violates the uncertainty bound around $x=1$.}
    \label{fig:illustrationUniformBoundsGPR}
\end{figure}

An important ingredient in SafeOpt-type algorithms are upper and lower bounds on the unknown target function, and these bounds have to hold uniformly both in time and input space. If we adopt a stochastic setup, then this can be formalized by finding upper and lower bounds such that for a given user-specified confidence $\delta\in(0,1)$, we have
\begin{equation*}
    \Pb\left[\ell_t(x) \leq f(x) \leq u_t(x), \: \forall x \in \setDec, t\geq 1 \right] \geq 1-\delta.
\end{equation*}
The probability is with respect to the data generating process, not the target function $f$ which is fixed. 
In GP regression, the posterior mean $\mu_t$ can be interpreted as a nominal estimate of $f$, and the posterior variance $\sigma_t^2$ as a measure of uncertainty of this estimate. However, using the posterior variance to build upper and lower bounds in the SafeOpt setting is not straightforward.
First, the posterior variance is a \emph{pointwise} measure of uncertainty about the ground truth, but the upper and lower bounds have to hold \emph{uniformly} over the input set.
Furthermore, the bounds have to hold also uniformly in time.
Second, GP regression is by its nature a Bayesian method. However, the SafeOpt setting is a typical frequentist setup -- we have a fixed, but unknown ground truth, and we receive noisy information about this ground truth. In particular, any stochasticity enters only through the data-generating process (e.g., via random noise on the function values), and not via epistemic uncertainty, as is the case in the Bayesian setup.
This difficulty is well-known, cf. \citet{fiedler2021aaai}, and is particularly relevant in the context of robust control and related areas \citep{fiedler2021cdc}.
What we need are bounds $(\nu_t)_{t\geq 1}$, such that for a user-specified $\delta\in(0,1)$ we have
\begin{equation} \label{eq:uniformBound}
    \Pb\left[|f(x)-\mu_t(x)| \leq \nu_t(x) \: \forall x\in \setDec, t\geq 1 \right] \geq 1-\delta.
\end{equation}
The bounding function $\nu_t$ must only depend on data collected up to time $t$, as well as reasonable prior knowledge about $f$ and the data-generating process, e.g., about the noise statistics.
Figure \ref{fig:illustrationUniformBoundsGPR} illustrates the situation.
All common bounds are of the form
\begin{equation} \label{eq:uniformBoundForm}
    \nu_t(x) = \beta_t \sigma_t(x),
\end{equation}
where $\beta_t\in\Rnn$ is some scaling factor.

Let $k$ be the covariance function used in GP regression, and denote by $H_k$ the unique RKHS having $k$ as its reproducing kernel.
Assume that the ground truth is contained in this RKHS, i.e., $f\in H_k$.
Let $\mathbb{F}=(\mathcal{F}_t)_{t \geq 0}$ be a filtration defined on the underlying probability space, and assume that the sequence of inputs $(x_t)_{t\geq 0}$ that is chosen by the algorithm is adapted to $\mathbb{F}$.\footnote{Of course, this entails that $\setDec$ is a measurable space, which is not a problem in practice.}

The first bound of the form \eqref{eq:uniformBoundForm} has been introduced in the seminal work \citet{Srinivas2010}, cf. their Theorem 6,
which holds in the case of bounded noise.
This is also the bound that has been used in the original SafeOpt paper \citet{sui2015safe}.
At the moment, the most commonly used uncertainty bound in the analysis of SafeOpt-type algorithms is \citet[Theorem~2]{Chowdhury2017}. 
Assume that $(\epsilon_t)_t$ is a martingale difference sequence that is conditionally $R$-subgaussian w.r.t. $\mathbb{F}$ for some $R\in\Rnn$, i.e., for all $t\geq 1$
\begin{equation*}
    \E\left[\exp(\nu \epsilon_t) \mid \mathcal{F}_t \right] \leq \exp\left(\frac{R^2\nu^2}{2}\right) \: \Pb\text{-a.s.} \quad \forall \nu\in\R.
\end{equation*}
Additionally, assume that $\lambda>1$, or $\lambda \geq 1$ and the covariance function $k$ is positive definite, where $\lambda$ is the nominal noise variance used in GP regression.
Under these conditions, the uncertainty bound \eqref{eq:uniformBound} holds with
\begin{equation} \label{eq:betaChowdhury}
     \beta_t = \|f\|_k + R\sqrt{2(\gamma_{t-1}+1+\log(1/\delta))}
\end{equation}
in \eqref{eq:uniformBoundForm}, where $\gamma_t$ is the \emph{maximum information gain} after $t$ rounds, cf. \citet{Srinivas2010} for a thorough discussion of this quantity.
In contrast to \citet[Theorem~6]{Srinivas2010}, this bound allows subgaussian noise (including bounded noise and normal noise) and involves only fairly small numerical constants. 
However, it still requires the maximum information gain or an upper bound thereof, which can be difficult to work with in practice, and it introduces some conservatism.

Motivated by these shortcomings, \cite{fiedler2021aaai} proposed a data-dependent scaling factor in \eqref{eq:uniformBoundForm}, based on \cite[Theorem~2]{Chowdhury2017}. Assume the same setting as this latter result, and that the covariance function $k$ is positive definite, then we can set
\begin{equation} \label{eq:betaFiedler}
    \beta_t = \|f\|_k + \frac{R}{\sqrt{\lambda}}\sqrt{\ln\left( \det(\bar{\lambda}/\lambda \K_t + \bar{\lambda} \Ident_t) \right) - 2\ln(\delta)},
\end{equation}
where we defined $\bar{\lambda}=\max\{ 1, \lambda \}$, and $\lambda$ is again the nominal noise variance used in GP regression, which corresponds to the regularization parameter in kernel ridge regression.
This bound does not involve the maximum information gain anymore, and numerical experiments demonstrate that the resulting uncertainty bounds are often not significantly larger than common heuristics, cf. \cite{fiedler2021aaai}. 
In fact, the bounds are small enough so that they can be used in algorithms, e.g. \cite{fiedler2021cdc}.

Finally, from the results in the doctoral thesis \cite{abbasi2012online}, which was published in 2012, an uncertain bound can be deduced that is superior to \cite[Theorem~2]{Chowdhury2017}, and therefore also improves over \eqref{eq:betaFiedler}. %
Consider the same setting as introduced above. Combining Theorem 3.11 with Remark 3.13 in \cite{abbasi2012online}, we find that for all $\delta\in(0,1)$ we can set
\begin{equation} \label{eq:betaAY}
    \beta_t = \|f\|_k + \frac{R}{\sqrt{\lambda}}\sqrt{2\ln\left(\frac{1}{\delta} \det\left(\Ident_t + \frac{1}{\lambda} \K_t \right) \right)}
\end{equation}
in \eqref{eq:uniformBoundForm}. This bound can be easily evaluated, just as \eqref{eq:betaFiedler}, though it seems that \eqref{eq:betaAY} has not appeared explicitly before.
Interestingly, this result appears to have been only infrequently used in the machine learning community, and has been rediscovered recently for example in \cite{whitehouse2023improved}. Furthermore, observe that for $0<\lambda<1$ and $k$ positive definite,
\begin{align*}
    \frac{R}{\sqrt{\lambda}}\sqrt{2\ln\left(\frac{1}{\delta} \det\left(\Ident_t + \frac{1}{\lambda} \K_t \right) \right)}
    & = \frac{R}{\sqrt{\lambda}}\sqrt{\ln\left( \det(1/\lambda \K_t + \Ident_t) \right) - 2\ln(\delta)} \\
    & = \frac{R}{\sqrt{\lambda}}\sqrt{\ln\left( \det(\bar{\lambda}/\lambda \K_t + \bar{\lambda} \Ident_t) \right) - 2\ln(\delta)},
\end{align*}
so in this case \cite[Theorem~1]{fiedler2021aaai} reproduces the result from \cite{abbasi2012online}.
Additionally, since the only difference between \eqref{eq:betaFiedler} and \eqref{eq:betaAY} happens inside $\sqrt{\ln(\cdot)}$, any noticable difference between the two bounds will happen for $\lambda >> 1$, so any difference will be neglegible in practice.

\section{Problem setting and objectives} \label{sec:problemSetting}
We now formalize our problem setting, describe SafeOpt-type algorithms in detail, and specify our objectives for the remainder of this work.
We work in the setting introduced by the seminal paper \citet{sui2015safe}. 
Consider a nonempty set $\setDec$, the \emph{input set}, and a fixed, but unknown function $f: \setDec\rightarrow\R$, the \emph{target function} or \emph{ground truth}.
We are interested in an algorithm that finds the maximum of $f$ by iteratively querying the function. 
At time step $t \in \N$, such an algorithm chooses an input $x_t \in \setDec$ and receives a noisy function evaluation $y_t = f(x_t)+\epsilon_t$, where $\epsilon_t$ is additive measurement noise. 
As a safety constraint, all chosen inputs must have a function value above a given safety threshold $\safeLevel \in \R$, i.e., $f(x_t) \geq \safeLevel$ for all $t$. 
Furthermore, the algorithm should be sample-efficient, i.e., use as few function queries as possible to find an input with a high function value.
It is clear that some restriction on the function $f$ must be posed to make progress on this problem. 
Central to our developments is the next assumption.
\begin{assumption} \label{assump:lipschitz}
$\setDec$ is equipped with a metric $d: \setDec \times \setDec \rightarrow \Rnn$.
Additionally, $f$ is $L$-Lipschitz continuous, where $L\in\Rnn$ is a known Lipschitz constant. 
\end{assumption}
The second assumptions means that for all $x,x^\prime \in \setDec$, we have $|f(x)-f(x^\prime)|\leq L \, d(x,x^\prime)$.
From now on, we work under Assumption \ref{assump:lipschitz}.
Furthermore, we assume that we have access to a non-empty set of known safe inputs $S_0 \subseteq \setDec$, i.e., for all $x \in S_0$ we have $f(x)\geq \safeLevel$.
\begin{algorithm}[t]
\caption{Generic SafeOpt-type algorithm}
\label{alg:safeopt}
\begin{algorithmic}
\State Initialize model $\mathcal{M}_0$
\For{$t =1,2,\ldots$}
    \State Compute current model $\model_t$
    \State Compute $S_t$ from $\model_t$
    \State Compute $\setExp_t$ from $S_t$ and $\model_t$
    \State Compute $\setMax_t$ from $S_t$ and $\model_t$
    \State $x_t \gets \argmax_{x \in \setExp_t \cup \setMax_t} w_t(x)$
    \State Query function with $x_t$, receive $y_t = f(x_t) + \epsilon_t$
    \State Update model with $(x_t,y_t)$
\EndFor
\end{algorithmic}
\end{algorithm}
The SafeOpt algorithm and its derivatives use an iteratively updated model $\model_t$ that provides estimated upper and lower bounds $u_t$ and $\ell_t$ on $f$, i.e., with a certain confidence it holds that $\ell_t(x)\leq f(x) \leq u_t(x)$ for all $x \in \setDec$ and all $t\geq 1$.
These bounds are also used to provide a measure of uncertainty defined as $w_t(x)=u_t(x)-\ell_t(x)$.
In each step $t\geq 1$, the previous model $\model_{t-1}$ together with the Lipschitz assumption is used to determine a new set $S_t\subseteq D$ of safe inputs, starting from the initial safe set $S_0$. 
Subsequently, a set $\setMax_t \subseteq S_t$ of potential maximizers of the target function, and a set $\setExp_t \subseteq S_t$ of potential expanders is computed. The latter contains inputs that are likely to lead to new safe inputs upon query.
Finally, the target function is queried at the input $x_t = \argmax_{x \in \setExp_t \cup \setMax_t} w_t(x)$, a noisy function value $y_t$ is received, and the model $\model_{t-1}$ is updated with the data point $(x_t,y_t)$. 
Pseudocode for a generic version of SafeOpt is provided by Algorithm \ref{alg:safeopt}. Different variants of SafeOpt result from different choices of models and computations of $S_t$, $\setMax_t$ and $\setExp_t$.

\begin{figure}[t]
    \includegraphics[width=\textwidth]{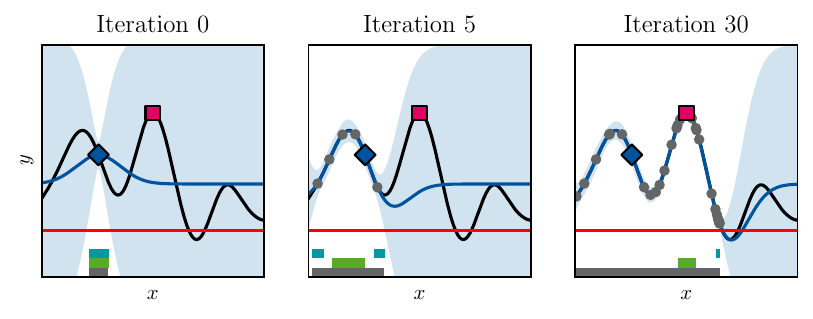}
    \caption{Illustration of SafeOpt. The safe set (gray bar), expanders (green bar), and maximizers (blue bar), which are derived from the current GP model (solid blue line is the posterior mean, shaded blue area are the uncertainty sets), are used to find the safely reachable optimum (red box).
    In each iteration, the next input is chosen from the union of the current expanders and maximizers (a subset of the safe set) by maximizing the acquisition function.}
    \label{fig:safeOptIllustration}
\end{figure}

To the best of our knowledge, in all SafeOpt-type algorithms, the unknown ground truth $f$ is modeled as a GP.
In order to compute appropriate upper and lower bounds, it is assumed that an appropriate scaling factor $\beta_t$ is available, cf. \eqref{eq:uniformBoundForm}.
For each time step $t$, define $C_t(x)=C_{t-1} \cap Q_t(x)$, where $Q_t(x)=[\mu_{t-1}(x) \pm \beta_t \: \sigma_{t-1}(x)]$,
and, starting with $Q_0(x)=\R$ for all $x \in \setDec$, $C_0(x)=[\safeLevel,\infty)$ for all $x \in S_0$ and $C_0(x)=\R$ for $x \in \setDec \setminus S_0$.
The corresponding estimated bounds are given by $u_t(x)=\max C_t(x)$ and $\ell_t(x)=\min C_t(x)$, respectively.
In the original SafeOpt algorithm from \cite{sui2015safe}, for each step $t \geq 1$, the new safe sets are calculated by
\begin{equation} \label{eq:safeopt:safeset}
    S_t=\bigcup_{x \in S_{t-1}}\{x' \in \setDec \, | \, \ell_t(x) - L \, d(x,x') \geq \safeLevel \},
\end{equation}
the potential maximizers are given by
\begin{equation} \label{eq:safeopt:maximizers}
    M_t = \{ x \in S_t \mid u_t(x) \geq \max_{x_S \in S_t} \ell_t(x_S) \},
\end{equation}
and the potential expanders by
\begin{equation} \label{eq:safeopt:expanders}
    G_t = \{ x_S \in S_t \mid \exists x \in \setDec \setminus S_t: \: u_t(x_S) - Ld(x_S,x) \geq \safeLevel  \}.
\end{equation}
The resulting algorithm is illustrated in Figure \ref{fig:safeOptIllustration}.
A formal description of the algorithm using pseudocode is provided by Algorithm \ref{alg:safeOptSpecific}.
\begin{algorithm}[t]
\caption{SafeOpt}
\label{alg:safeOptSpecific}
\begin{algorithmic}[1]
\Require Lipschitz constant $L$, algorithm to compute $\beta_t$, initial safe set $S_0$, safety threshold $h$
\State $Q_0(x) \gets \mathbb{R}$ for all $x \in D$ \Comment{Initialization of uncertainty sets}
\State $C_0(x) \gets [h,\infty)$ for all $x \in S_0$
\State $C_0(x) \gets \mathbb{R}$ for all $x \in D \setminus S_0$
\For{$t =1,2,\ldots$}
    \State $C_t(x) \gets C_{t-1}(x) \cap Q_{t-1}(x)$ for all $x \in D$ \Comment{Compute upper and lower bounds for current iteration}
    \State $\ell_t(x) \gets \min C_t(x)$, $u_t(x) \gets \max C_t(x)$ for all $x \in D$
    \If{$t > 1$} \Comment{Compute new safe set}
        \State $S_t = S_{t-1} \cup \{ x \in D \mid \exists x_s \in S_{t-1}: \: \ell_t(x_s) - Ld(x_s, x) \geq h \}$
    \Else
        \State $S_1 = S_0$
    \EndIf
    \State $G_t \gets \{ x \in S_t \mid \exists x^\prime \in D \setminus S_t: \: u_t(x) - Ld(x,x^\prime) \geq h  \}$ \Comment{Compute set of potential expanders}
    \State $M_t = \{ x \in S_t \mid u_t(x) \geq \max_{x_S \in S_t} \ell_t(x_S) \}$ \Comment{Compute set of potential maximizers}
    \State $x_t \gets \argmax_{x \in G_t \cup M_t} w_t(x)$ \Comment{Determine next input}
    \State Query function with $x_t$, receive $y_t = f(x_t) + \epsilon_t$
    \State Update GP with new data point $(x_t,y_t)$, resulting in mean $\mu_t$ and $\sigma_t$
    \State Compute updated $\beta_t$
    \State $Q_t(x)=[\mu_t(x) - \beta_t \: \sigma_t(x), \mu_t(x) + \beta_t \: \sigma_t(x)]$ for all $x \in D$ 
\EndFor
\end{algorithmic}
\end{algorithm}
In some popular variants of the SafeOpt algorithm, no Lipschitz bound is used in the computation of the safe sets \citep{berkenkamp2016safe}.
However, since the knowledge of such a Lipschitz bound is additional knowledge which should be used by the algorithm, and we strongly rely on this assumption from Section \ref{sec:losbo} onwards, we do not consider these algorithmic variants in the present work.

Our primary objective is to investigate and improve practically relevant safety aspects of SafeOpt-type algorithms.
We will consider three specific objectives.
\begin{description}
    \item[(O1)] While SafeOpt and related algorithms come with precise theoretical safety guarantees, all known implementations so far use heuristics instead of theoretically sound uncertainty bounds.
    In general, these heuristics invalidate all theoretical safety guarantees, which raises the question whether this leads to practical safety problems.
    \item[(O2)] Without heuristics, safety guarantees for SafeOpt-type algorithms rely on the appropriateness of the assumptions used for the algorithmic setup.
    Since SafeOpt and its variants are used for scenarios with stringent safety requirements, where no failure should occur with high probability, and no tuning phase is allowed, it is important that all assumptions are reasonable and verifiable by users.
    The most delicate of these assumptions is the knowledge of an upper bound of the RKHS norm of the target function.
    This poses the question whether this the RKHS norm bound assumption is reasonable, and if not, how it can be replaced.
    \item[(O3)] In many relevant applications of safe BO, the input space is a continuous domain in moderately high dimension.
    Since SafeOpt-type algorithms rely on a discrete input space, they cannot be used in these settings.
    It is therefore important to devise alternatives that work in moderately high dimensions, while retaining safety guarantees.
\end{description}
The three aspects will be investigated in Sections \ref{sec:realBeta}, \ref{sec:losbo} and \ref{sec:highdim}, respectively, after discussing related work.
\section{Related work} \label{sec:relatedWork}
In the following, related work will be reviewed.
We will first provide an overview of safe BO methods, in particular those that are closely related to the present setting.
Since from Section \ref{sec:losbo} onwards, Lipschitz bounds will play a central role, we will also review previous work on methods relying on such an assumption.

\subsection{Safety in Bayesian Optimization}
We focus on safety in the context of Bayesian optimization (BO) \citep{Shahriari2015,garnett2023bayesian}.
The type of safety constraint considered in the present work has been introduced in the seminal paper \cite{sui2015safe}, which also proposed and analyzed the original SafeOpt algorithm.
A safety constraint of this type appears in many practically relevant application scenarios, for example, controller tuning \citep{berkenkamp2016safe,berkenkamp2023bayesian}, automatic tuning of medical therapies \citep{Sarikhani2021}, or safe robot learning \citep{baumann2021gosafe}.
In terms of algorithmic variations, a SafeOpt-variant not using a Lipschitz constant has been proposed \citep{berkenkamp2016safe}, which has been particularly popular in the robotics community.
While in the original SafeOpt algorithm from \citet{sui2015safe} safe exploration and function optimization over certified safe sets are interleaved, a two-stage variant has been introduced and analyzed in \citet{sui2018stagewise}.
Since SafeOpt relies on a discrete input space, and hence requires discretization in case of a continuous optimization problem, applying this algorithm to continuous problems in even moderate dimensions can be very challenging, cf. also Section \ref{sec:highdim}.
This motivated the introduction of a variant based on swarm-optimization \citep{duivenvoorden2017constrained}, albeit only with heuristic safety guarantees, as well as a variant tailored to dynamical problems in high dimensions \citep{sukhija2023gosafeopt}.
Furthermore, a method based on random projections has been proposed and analyzed in \citet{kirschner2019adaptive}, and applied to tuning of a particle accelerator \citep{kirschner2019bayesian}.
In terms of problem formulations, instead of just one safety constraint of the input function, multiple safety constraints can also be enforced, each encoded by an unknown constraint function \citep{berkenkamp2016bayesian,sui2018stagewise,berkenkamp2023bayesian}.
In many applications of SafeOpt-type algorithms, properties of a dynamical system should be optimized under safety constraints.
For example, in controller tuning with BO, the inputs correspond to parameters of a controller, and the performance of the controller is the target function to be optimized. 
The dynamic aspect can be included in the optimization algorithm and its safety mechanisms, for example, in order to use a backup controller \citep{baumann2021gosafe}.
Furthermore, while formally SafeOpt-type algorithms are BO algorithms having input constraints, this is different from \emph{constrained BO} as considered for example in \citet{hernandezlobato2016general}.
In the latter type of BO, one is interested in finding good inputs (i.e., corresponding to a high objective function value) fulfilling the (usually unknown) constraints, but violation of the constraints during the optimization process is not considered problematic (though of course, it can be advantageous to avoid them).
In contrast to this, in safe BO the want to avoid constraints violations during the optimization process since these are considered problematic or even harmful.
For an in-depth discussion of this difference, and further connections between the two problem settings, we refer to the excellent survey \citet{kim2020safe}.

The motivation for SafeOpt-type algorithms are problems where safety violations are considered to be very costly.
In particular, one would like to avoid \emph{any} safety violation with very high probability.
For example, in the case of controller tuning a safety violation could correspond to damage of the controlled plant, or in robot policy tuning a safety violation would correspond to a literal crash of an actual robot, potentially destroying it.
Similarly, in a medical context a safety violation corresponds to patient harm and therefore has to be avoided with very high probability.
In short, SafeOpt-type algorithms are particularly interesting for \emph{hard safety requirements}, i.e., avoiding any safety constraint violation (with high probability).
This is in constrast to a whole spectrum of related, but different safe BO (and reinforcement learning) settings.
For example, in robot policy optimization, a crash of a robot might not lead to damage, but rather correspond to an unsuccessful experimental run.
A small number of crashes becomes then acceptable, since it might allow more aggressive exploration, which in turn leads to better policies after optimization.
SafeOpt-type algorithms, tailored to avoiding \emph{any} safety violations, are therefore not an optimal choice in this scenario, and one would rather use a crash-aware BO algorithm \citep{marco2021robot}.
Similarly, in the context of bandit algorithms \citep{slivkins2019introduction,lattimore2020bandit}, instead of avoiding a bad option at all costs, one might instead want to be \emph{conservative}, carefully exploring alternatives, starting from a default strategy.
This setting is formalized in the context of \emph{conservative bandits} \citep{wu2016conservative}, and again, is related but distinct from the SafeOpt setting.
Corresponding bandit formulations in the context of hard safety are also available \cite{amani2019linear}, where the focus is on regret bounds under safety constraints.
Finally, parallel to the situation of conservative bandits, one can consider cautious variants of BO \citep{frohlich2021cautious}.
Safety violations are not avoided under all costs, but rather the exploration tries to proceed cautiously.

\subsection{Lipschitz-based methods}
In order to address safety-related issues uncovered and discussed in Sections \ref{sec:heuristicsProblem} and \ref{sec:problemWithRKHSnormBound}, we will introduce algorithms based on a Lipschitz assumption. In particular, it will be assumed that the target function is Lipschitz-continuous with a known upper bound on the Lipschitz constant.
While regularity properties like Lipschitz (and the more general Hölder) continuity play an important role in the theory of statistical learning, in particular, nonparametric statistical estimation \citep{tsybakov2009nonparametric},
learning algorithms based on Lipschitz assumptions have received relatively little attention in the machine learning community.
One exception related to the present context are Lipschitz bandits \citep{kleinberg2019bandits},
and the original SafeOpt algorithm from \citet{sui2015safe}.
The situation is considerably different in the field of global optimization and the systems and control community, respectively.

In the former, Lipschitz continuity with a specific Lipschitz constant is a standard assumption, used in a variety of algorithms for (certified) global optimization, cf. \citet{hansen1992global,pinter1995global}, though usually a noise-free setting is assumed in this literature.
This problem has recently also received attention from the machine learning community \citep{malherbe2017global}.
Similarly, a specified Lipschitz constant is also used in the context of Lipschitz interpolation \citep{beliakov2006interpolation}.
Furthermore, closely related to our approach taken in Section \ref{sec:losbo}, a deterministic variant of SafeOpt has been considered in \citet{sergeyev2020safe}. However, the latter reference only works with functions on a compact interval, and does not use any BO technique.

In the systems and control community, Lipschitz assumptions have been used for a considerable amount of time, in particular, in the context of systems identification, where it has been explicitly introduced and popularized by \citet{milanese2004set}, though similar methods have been used before, e.g., \citet{cooper1995learning}. In particular, a known bound on the Lipschitz constant and on the size of additive noise is used to derive uncertainty bounds in the context of regression.
This approach has been further popularized and extended to the case of Hölder continuous functions by \citet{calliess2014conservative}, which is commonly called \emph{kinky inference} in the systems and control community.
A central assumption in this context is the knowledge of a \emph{concrete, numerical} upper bound on the Lipschitz constant of the target function.
This assumptions has a clear geometric and practical interpretation, namely a bounded rate of change of the target quantity. As such, it is for example related to the well-established field of sensitivity analysis \citep{daveiga2021basics}.

Approaches to estimate the Lipschitz constant of an unknown function are proposed both in the context of global optimization \citep{strongin1973convergence} as well as Lipschitz based regression methods, in particular, in the context of systems identification \citep{milanese2004set,novara2013direct,calliess2020lazily}, cf. \cite{huang2023sample} for an overview and very recent sample-complexity results.
We would like to stress that these approaches are not suitable for the present setting of hard safety constraints, since the estimation of a Lipschitz constant bound requires queries to the target function, which in turn already need to be safe, cf. also the discussion in Section \ref{sec:problemWithRKHSnormBound}.

The developments in Sections \ref{sec:losbo} and \ref{sec:highdim} combine kernel-based methods (here GP regression) with a Lipschitz assumption, in particular, to overcome the requirement of a known bound on the RKHS norm of the target function, cf. Section \ref{sec:problemWithRKHSnormBound} for details.
The problematic nature of an RKHS norm bound in the context of learning-based control has been recognized for a while \citep{lederer2019uniform,fiedler2022learning}.
In \citet{lederer2019uniform}, using probabilistic Lipschitz bounds together with a space discretization has been suggested to derive GP uncertainty bounds, however, this approach relies on a probabilistic setting, and is therefore not suitable in the context of SafeOpt-type algorithms.
The work \citet{fiedler2022learning} proposes the usage of geometric constraints as prior knowledge in the context of uncertainty sets for kernel-based regression, with Lipschitz constant bounds as a special case.
The resulting kernel machines, providing nominal predictions and smoothed uncertainty bounds adhering to the geometric constraints, are not necessary in our setting, and using more general geometric constraints than Lipschitz constant bounds might be a promising avenue in the context of SafeOpt-type algorithms, which we leave for future work.
Finally, combining kernel methods with Lipschitz assumptions is a natural approach, since it is well-known that there is a close connection between regularity properties of a kernel and Lipschitz continuity of functions in the RKHS generated by the kernel. 
For a thorough discussion of this aspect, we refer to \citet{fiedler2023lipschitz}.

\section{Frequentist uncertainty bounds and practical safety issues in SafeOpt} \label{sec:realBeta}
We now investigate practical safety implications of commonly used heuristics in the frequentist uncertainty bounds in SafeOpt-type algorithms, addressing objective (O1).
In Section \ref{sec:heuristicsProblem}, we discuss why these heuristics are problematic and demonstrate safety issues using numerical experiments.
To overcome these problems, in Section \ref{sec:realBetaSafeOpt} we propose to use state-of-the-art frequentist uncertainty bounds in the actual algorithm.
\subsection{Practical safety issues in SafeOpt} \label{sec:heuristicsProblem}
Safety in SafeOpt-type algorithms is ensured by restricting query inputs to safe sets, which in turn are computed using frequentist uncertainty bounds,
in particular, in the form \eqref{eq:uniformBoundForm} using \eqref{eq:betaChowdhury}. 
However, these bounds are often too conservative for algorithmic use, leading to existing implementations adopting heuristic choices for $\beta_t$, for example, $\beta_t \equiv 2$ in \citet{berkenkamp2016safe,Turchetta2016}, $\beta_t \equiv 3$ in \citet{Helwa2019,baumann2021gosafe} or $\beta_t \equiv 4$ in \citet{Sukhija2022}.
Using such heuristics instead of evaluating $\beta_t$ \textbf{invalidates all theoretical safety guarantees}.
Choosing some $\beta_t$ can be a useful heuristic in practice, as demonstrated by the reported success of SafeOpt-type algorithms \citep{berkenkamp2016safe,baumann2021gosafe,sukhija2023gosafeopt}.
However, it should be stressed that in the setting of SafeOpt as outlined in Section \ref{sec:problemSetting}, the learning algorithm has to fulfill a \textbf{hard safety constraint} -- namely, that no unsafe inputs are queried by the algorithm at all (potentially only with high probability). In particular, no burn-in or tuning phase for $\beta_t$ is allowed.
Furthermore, not only are the theoretical safety guarantees invalidated by such heuristics, they can actually lead to safety violations.

First, we demonstrate empirically that a simple heuristic like setting $\beta_t \equiv 2$ can lead to a significant proportion of bound violations.
To do so, we follow the general approach from \cite{fiedler2021aaai}.
We generate randomly 100 RKHS functions on $[-2,2]$ with RKHS norm \expm{10}, using the squared exponential kernel with length scale \expm{$0.2/\sqrt{2}$}.
For this, the orthonormal basis (ONB) of the corresponding RKHS as described in \citet[Section~4.4]{SC08} is utilized, by selecting some of these basis functions, and combining them in a weighted sum, using randomly generated weights.
For each of the resulting functions, we generate 10000 independent data sets, by uniformly sampling \expm{$100$} inputs from \expm{$[0,1]$}, evaluating the RKHS function on these inputs, and then adding i.i.d. normal noise with variance \expm{$0.01$}.
For each data set, we apply GP regression with a zero mean prior and the SE kernel as covariance function, using the same length scale as for the generation of the target functions.
Finally, we use the uncertainty set \eqref{eq:uniformBoundForm} with $\beta_t\equiv 2$, and check on a fine grid on $[-2,2]$ whether the target function from the RKHS is completely contained in this uncertainty set. 
It turns out that \expm{2727}$\pm$\expm{3882} (average $\pm$ SD) of these runs (all 10000 repetitions for all 100 functions) lead to a bound violation, which is a rather sizeable proportion.

Second, these bound violation can indeed lead to safety violations when running SafeOpt.
To demonstrate this, we have generated a RKHS function $f$ (same approach as above), and defined the safety threshold $\safeLevel$ by setting \expm{$h=\hat\mu(f)-0.2\hat{SD}(f)$}, where $\hat\mu(f)$ and $\hat{SD}(f)$ are the empirical mean and standard deviation of the test function $f$ evaluated on a fine grid of the input space.
Furthermore, we evaluate $|f'|$ on a fine grid, take the maximum and multiply it by 1.1 to find a (slightly conservative) upper bound of the Lipschitz constant of $f$.
We then run SafeOpt on this function 10000 times from a random safe initial state, using again i.i.d. additive normal noise with variance \expm{0.01}.
This leads to \expm{2862} (out of 10000) runs with safety violations (cf. Table \ref{tab:SafetyPerfomanceSafeOpt}), which is certainly unacceptable for most application scenarios of SafeOpt-type algorithms.

These experiments illustrate that using heuristics in SafeOpt can be very problematic.
On the one hand, even in the relative benign setting used above, both uncertainty bound and safety violations do occur.
On the other hand, in the application scenario of SafeOpt-type algorithms, there is no possibility to tune the heuristic scaling factor, since it is the primary mechanism for safety. 
Dispensing with the need for such heuristics, and retaining safety guarantees both in theory and practice, is therefore the primary motivation for this work.

\begin{remark}
The assumption $f\in H_k$ is standard in the BO literature \citep{abbasi2012online, Srinivas2010,Chowdhury2017,whitehouse2023improved}, where $k$ is a kernel that is used as the covariance function in GP regression.
In general, if functions from $H_k$ are to be used as models of the underlying target function, the RKHS $H_k$ has to be sufficiently rich.
In many cases, as a qualitative assumption, this is rather mild. 
For example, if $D$ is a compact metric space, then a large variety of \emph{universal kernels} are available, which are kernels with an RKHS that is dense (w.r.t. the supremum norm) in the set of all continuous functions on $D$ \citep[Section~4.6]{SC08}.
However, the assumption $f\in H_k$ is much more stringent and delicate.
In particular, even if the target function is contained in an RKHS generated by a kernel from the same class as the one used as a covariance (e.g., a squared exponential or Matern kernel), this might not be enough because there can be a mismatch of hyperparameters.
For the case of squared exponential kernels, a complete characterization of the inclusions of RKHSs w.r.t. different hyperparameters is described in \citet[Section~4.4]{SC08}, revealing that this situation can indeed occur.
While there is work on covariance function misspecification in GP regression \citep{wynne2021convergence}\footnote{Even likelihood misspecification.} and GP-based BO \citep{berkenkamp2019unknown,bogunovic2021misspecified}, these works do not help in the present situation, because of the form of the resulting guarantees in the former case, and that the schemes in the latter case might sample unsafe inputs in the latter case.
The situation is additionally complicated since in practice, the hyperparameters are adapted during BO \citep{garnett2023bayesian}, theoretical guarantees of which are only slowly emerging \citep{teckentrup2020convergence,karvonen2020maximum,karvonen2023maximum}.
In the following, we work (unless noted otherwise) with the assumption $f\in H_k$, and do not consider hyperparameter adaption during the optimization process.
This is justified by the following two aspects. First, this is a common approach in the BO literature, including safe BO \citep{sui2015safe,sui2018stagewise}.
Second, and more importantly, the safety-related issues that we are about to discuss are independent of this issue (though of course, they are an additional problematic factor).
\end{remark}

\subsection{Real-$\beta$-SafeOpt} \label{sec:realBetaSafeOpt}
As a first step, we propose to use modern uncertainty bounds in SafeOpt that can be computed numerically, avoiding the replacement with unreliable heuristics.
For this purpose, we investigate the original SafeOpt algorithm with $\beta_t$ from \eqref{eq:betaAY}, which is very close to \eqref{eq:betaFiedler}.
To clearly distinguish this variant of SafeOpt from previous work, we call it \emph{Real-$\beta$-SafeOpt}, emphasizing that we use a \emph{theoretically sound choice of $\beta_t$}.
The resulting algorithm is again described by Algorithm \ref{alg:safeOptSpecific}, using \eqref{eq:betaAY} to compute $\beta_t$.
The bound \eqref{eq:betaAY} requires computing the determinant, which can be computationally expensive, but typical applications of SafeOpt and related algorithms allow only few evaluations, so that this does not pose an issue.
Furthermore, the additive noise needs to be a conditionally $R$-subgaussian (martingale-difference) sequence with a (known upper bound on) $R$.
This assumption is standard, and in many cases harmless. It also has a clear interpretation.
Finally, for a frequentist uncertainty bound, we also need the next assumption.
\begin{assumption} \label{assump:rkhsNormBound}
Some $B\in\Rnn$ is known with $\|f\|_k\leq B$.
\end{assumption}
Combining these ingredients allows us to compute the bound \eqref{eq:betaAY}, so Real-$\beta$-SafeOpt can actually be implemented.
To illustrate the advantages of Real-$\beta$-SafeOpt, we run it on the same function $f$ as before (cf. Section \ref{sec:heuristicsProblem}), and set  \expm{$\delta=0.01$}, as well as the true RKHS norm in \eqref{eq:betaAY}.
Running this experiment results in
\expm{0} failures (cf. Table \ref{tab:SafetyPerfomanceSafeOpt}), obviously well within the $\delta$ range required.

An obvious relevant question is how Real-$\beta$-SafeOpt compares in performance with previous SafeOpt-variant relying on heuristics.
Unfortunately, a meaningful comparison is impossible since the two algorithmic variants address different problems.
If for the latter the heuristic constant $\beta$ is determined by trial-and-error, and then SafeOpt is run with this constant which leads to no or almost no safety violation, then one ends up essentially with a different algorithm overall.
If the heuristic constant $\beta$ is chosen arbitrarily or based on experience with previous usages of SafeOpt-type algorithms, then one leaves the original setting of hard safety constraints, and instead ends up with a form of cautious BO.
In contrast, Real-$\beta$-SafeOpt tries to stay within the original setting of SafeOpt requiring hard safety constraints.
To investigate how Real-$\beta$-SafeOpt behaves typically, in Section \ref{sec:losboExperiments} we will perform a careful empirical evaluation of this algorithm variant.

\section{Lipschitz-only Safe Bayesian Optimization (LoSBO)} \label{sec:losbo}
As discussed above, the safety of SafeOpt-type algorithms should only depend on reasonable assumptions that can be verified and interpreted by the algorithm's user. 
In this section, we address objective (O2).
In particular, we investigate the central Assumption \ref{assump:rkhsNormBound} of a known upper bound on the RKHS norm of the target function, finding that this is a problematic assumption in practice.
To overcome this issue, we propose Lipschitz-only Safe Bayesian Optimization (LoSBO) as a solution to this problem, and perform extensive numerical experiments, comparing LoSBO against Real-$\beta$-SafeOpt.
\subsection{Practical problems with the RKHS norm bound} \label{sec:problemWithRKHSnormBound}
A central ingredient in the Real-$\beta$-SafeOpt algorithm is an upper bound on the RKHS norm of the target function.
In particular, the safety and exploration guarantees inherited from \citet{sui2015safe} hinge on the knowledge of such an upper bound.
Unfortunately, while the RKHS norm is very well understood from a theoretical point of view, it is unclear how to derive such a bound in practice.
In the following, we give an overview of known characterizations and representations of the RKHS norm, and discuss why these result seem to be not suitable for this task.

For an arbitrary kernel, one can use discretization-based variational characterizations of the RKHS norm (and RKHS functions), for example,
by maximization over a family of lower bounds on the RKHS norm \citep[Section~B]{fiedler2023kernel}, \citep[Chapter~I]{atteia1992hilbertian},
by minimization over certain bounds on function values at finitely many inputs \citep[Theorem~A.2.6]{okutmucstur2005reproducing},
by minimization over finite interpolation problems \citep[Theorem~3.11]{PR16},
or by minimization over certain matrix inequalities \citep[Theorem~3.11]{PR16}.
For separable RKHSs, the RKHS norm can be expressed using a sampling expansion \citep{korezlioglu1968reproducing}, 
or as the limit of norms of RKHSs over finite inputs \citep[Lemma~4.6]{lukic2001stochastic}.
On the one hand, all of these variational problems have an explicit form and they work for \emph{any} kernel (any kernel with separable RKHS, respectively).
However, it is not clear at all how to relate these representations to common properties of functions that might be used as reliable prior knowledge to derive upper bounds on the RKHS norm.
Furthermore, in general these variational problems cannot be used in numerical methods to estimate upper bounds on the RKHS norm, but only lower bounds, though they might be used in heuristics for estimating bounds \citep{tokmak2023automatic}.
Additionally, since these characterizations are based on discretizations of a given RKHS function, in particular, using the exact function values, they are not suitable in the present setting of unknown target functions accessible only through noisy evaluations.

If one considers more specific classes of kernels, other characterizations of the RKHS norm become available.
For example, continuous kernels on a compact metric space equipped with a measure having full support (often called a Mercer kernel in this context) allow a description of the RKHS norm as a weighted $\ell_2$-norm \citep[Section~4.5]{SC08}, based on Mercer's theorem.
This has a clear interpretation in the context of kernel methods, in particular, giving insight into the regularization behavior of the RKHS norm in optimization problems in kernel machines \cite[Section~5.8]{hastie2009elements},
which in turn can be used to derive learning rates for various statistical learning problems \citep{steinwart2009optimal}.
More general forms of Mercer's theorem are available \citep{steinwart2012mercer}, which in turn lead to improved learning theory results \citep{fischer2020sobolev}.
While the RKHS norm representation for Mercer kernels is an important tool for statistical learning theory and provides intuition about the regularization behavior, it is again unclear how it can be used to derive \emph{quantitative} RKHS norm bounds.
Expressing the RKHS norm for Mercer kernels as a weighted $\ell_2$-norm provides valuable \emph{qualitative} intuition about the corresponding RKHS norm, but we are not aware of any practically relevant example where this has been used to translate realistic prior knowledge into a concrete upper bound on the RKHS norm.

Similarly, for sufficiently regular translation-invariant kernels, one can express the RKHS norm as a weighted integral over the Fourier transform of RKHS functions \citep[Theorem~10.12]{wendland2004scattered}. This allows an intuitive interpretation of the RKHS norm as a generalized energy, penalizing high-frequency behavior of RKHS functions (as determined by the Fourier transforms of the kernel).
Several important function spaces are related to RKHSs, for example certain Sobolev spaces \citep[Chapter~10]{wendland2004scattered} or Fock spaces \citep[Section~4.4]{SC08}, having again their own representations of the RKHS norm (potentially after some embedding).
Again, all of these representations allow to build intuition about the RKHS norm, and are important theoretical tools, but it is unclear how this can be used to derive practically useful quantitative upper bounds on the RKHS norm.

To summarize, while an extensive body of characterization and representation results for the RKHS norm is available, these results appear to be unsuitable to derive upper bounds on the RKHS norm. 
In particular, to the best of our knowledge, it is not possible at the moment to derive concrete numerical upper bounds on the RKHS norm from realistic assumptions in non-trivial cases.
The central difficulty here is that one needs a concrete \emph{numerical upper bound} on the RKHS norm for algorithms like Real-$\beta$-SafeOpt.
This issue is known, in particular, in the learning-based control community \citep{fiedler2022learning}, but not addressed yet in the context of safe BO.
One reason might be that this problem does not appear in many other kernel-based learning scenarios. For example, in order to derive learning rates for SVMs and related kernel machines, membership of the target function\footnote{In the standard statistical learning theory setup, there is no notion of a target function as in the context of safe BO. Instead, the learning problem is described using a loss function. However, for example in the context of regression with the squared-error loss, the conditional expectation (the regression function) takes the role of our target function in the explanation above.} in an appropriate RKHS is enough (or in a function space that in a suitable sense can be approximated by an RKHS), and (an upper bound of) the RKHS norm of the target function is \emph{not} needed by the kernel-based learning algorithm.

The preceding discussion has immediate consequences for SafeOpt-type algorithms.
These algorithms are meant for hard safety settings, i.e., scenarios where \emph{any} safety violation is very costly and has to be avoided.
In particular, in these scenarios one cannot have a tuning phase before the actual optimization run, where algorithmic parameters are set - the safety requirements hold from the start. 
For SafeOpt-type algorithms this means that all algorithmic parameters have to be set beforehand, and these parameters need to ensure the safety guarantees and lead to satisfying exploration behaviour.
However, this entails in particular an upper bound on the RKHS norm of the target function, which at the moment appears to be impossible to derive from reasonable prior knowledge in practically relevant scenarios.

Furthermore, using an invalid RKHS upper norm bound can indeed easily lead to safety violations.
In order to illustrate this, we run SafeOpt on the same function $f$ as before (cf. Section \ref{sec:heuristicsProblem}), and set \expm{$\delta=0.01$}, but this time, we use a misspecified RKHS norm of \expm{2.5} in \eqref{eq:betaAY}.
Running this experiment now results in \expm{1338} failure runs out of \expm{10000}, cf. Table \ref{tab:SafetyPerfomanceSafeOpt}, which is much more than what would be expected from a safety probability of $1-\delta=0.99$.
For a summary of this experiment, see again Table \ref{tab:SafetyPerfomanceSafeOpt}.

Finally, simply using a very conservative upper bound on the RKHS norm is not a viable strategy to overcome this problem.
On the one hand, a severe overestimation of the RKHS norm leads to very large and overly conservative uncertainty bounds, which in turn leads to performance issues.
In particular, since the uncertainty bounds are used to determine the safety sets in SafeOpt-type algorithms, a supposed RKHS upper norm bound that is too conservative can result in the algorithm ``getting stuck'', i.e., no more exploration is possible.
On the other hand, it is not even clear what ``very conservative'' means in the present context.
Recalling the discussion of the RKHS norm from above, while an extensive body of theory and strong qualitative intuitions on this objects are available, the lack of concrete, quantitative bounds amenable to numerical evaluation makes it very difficult for users of BO to judge what a conservative estimate of the RKHS norm in a given situation could be.

We argue that as a consequence, any SafeOpt-like algorithm \emph{should not depend on the knowledge of an upper bound on the RKHS norm of the target function for safety},
at least in the setting of hard safety requirements where \emph{any} failure should be avoided (with high probability).
More generally, in order to guarantee safety, we should only use assumptions that are judge as reliable and reasonable by practitioners.
In particular, all assumptions that are used for safety should have a clear interpretation for practitioners and a clear connection to established prior knowledge in the application area of the safe BO algorithm.
In the end, it is up to the user of safe BO to decide which assumptions used in the safety mechanism can be considered as reliable.
\subsection{Describing LoSBO and its safety guarantees} \label{sec:introducingLoSBO}
Motivated by the popularity of SafeOpt, which combines GPs with a Lipschitz assumption, and the extensive experience of the systems and control community with Lipschitz bounds and bounded noise (cf. Section \ref{sec:relatedWork}), we propose to use \emph{an upper bound on the Lipschitz constant of the target function and a known noise bound} as the ingredients for safety in BO.
More precisely, we propose Lipschitz continuity of the target function and a known upper bound on the Lipschitz constant, together with bounded measurement noise and a known upper bound on the magnitude of the noise, as core assumptions for safety.
Both of these assumptions fulfill the desiderata outlined above. 
First, they have a clear interpretation: A known upper bound on the Lipschitz constant corresponds to a slope constraint on the target function, and bounded noise is self-explanatory.
Second, they are easily related to established prior knowledge: A known upper bound on the Lipschitz constant corresponds to an a priori bound on the rate of change of a function, i.e., it is related to the sensitivity of the underlying problem, and a known bound on the magnitude of the noise means that the strength of the noise generating mechanism is known.

The key idea is to make sure the safety mechanism works reliably independent of the (statistical) exploration mechanism.
In a generic SafeOpt algorithm, safety is guaranteed by ensuring that the safe sets $S_t$ contain only safe inputs (potentially only up to high probability), i.e., requiring that $f(x)\geq \safeLevel$ for all $x\in S_t$.
Once this property is fulfilled, the rest of the algorithm cannot violate the safety constraints anymore.
Based on the preceding discussion, the construction of the safe set should only rely on the Lipschitz and noise bound.
As is well-known, these two assumption allow the construction of lower bounds on the function \citep{milanese2004set}, and the corresponding safe sets should therefore be defined for all $t\geq 1$ as
\begin{equation} \label{eq:losbo:safeset}
    S_t = S_{t-1} \cup \{ x \in D \mid y_{t-1} - E - Ld(x_{t-1}, x) \geq h \},
\end{equation}
where $L\in\Rnn$ is a bound on the Lipschitz constant of the unknown target function, $E\in\Rnn$ a bound on the magnitude of the noise, and $S_0$ is the initial safe set.
We propose to use this variant of the safe set, and leave the rest of the generic SafeOpt algorithm unchanged, which leads to an algorithm we call Lipschitz-only Safe Bayesian Optimization (LoSBO).
It fulfills the following safety guarantee.
\begin{proposition} \label{prop:losbo:safety}
Let $f: D \rightarrow \mathbb{R}$ be an $L$-Lipschitz function. Assume that $|\epsilon_t| \leq E$ for all $t \geq 1$ and let $\emptyset \not = S_0 \subseteq D$ such that $f(x) \geq h$ for all $x \in S_0$. For any choice of the scaling factors $\beta_t>0$, running the LoSBO algorithm leads to a sequence of only safe inputs, i.e., we have $f(x_t) \geq h$ for all $t \geq 1$.
\end{proposition}
\begin{proof}
It is enough to show that $\forall t \geq 0$ and $x \in S_t$, we have $f(x)\geq h$. Induction on $t$: 
For $t=0$, this follows by assumption. For $t \geq 1$, let $x \in S_t = S_{t-1} \cup \{ x \in D \mid y_{t-1} - E - Ld(x_{t-1}, x) \geq h \}$.
If $x \in S_{t-1}$, then $f(x)\geq h$ follows from the induction hypothesis. Otherwise we have
\begin{align*}
    f(x) & = f(x_t) + \epsilon_t - \epsilon_t + f(x) - f(x_t) \geq y_t - E - Ld(x_t,x) \geq h,
\end{align*}
where we used the $L$-Lipschitz continuity of $f$ and the noise bound $|\epsilon_t|\leq E$ in the first inequality,
and the definition of $S_t$ in the second inequality.
\end{proof}
The argument in the proof above is well-known, for example, in the systems identification literature, and the result bounds fulfill even certain optimality properties \citep{milanese2004set}. %
Finally, we would like to stress that the safety guarantee of LoSBO, as formalized in Proposition \ref{prop:losbo:safety}, is \emph{deterministic}, i.e., it always holds and not only with high probability.
This type of safety is often preferred in the context of control and robotics \citep{hewing2020learning,brunke2022safe}.
\begin{remark} \label{remark:generalizations}
Inspecting the proof of the preceding result reveals that considerably more general (and weaker) assumptions can be used with the same argument.
\begin{enumerate}
    \item Instead of a fixed noise bound $E$, one can mutatis mutandis use asymmetric, time-varying and heteroscedastic noise. 
    Formally, one can assume that two functions $E_\ell,E_u: \setDec \times \N \rightarrow\Rnn$ exist, such that for all $t\geq 1$ and $x\in\setDec$, it holds that $E_\ell(x,t)\leq \epsilon_t \leq E_u(x,t)$, if $x$ is the input used at time $t$.
    \item Instead of Lipschitz continuity, one can assume that there exists a continuous and strictly increasing function $\phi:\Rnn\rightarrow\Rnn$ with $\phi(0)=0$, such that for all $x,x'\in\setDec$ it holds that $f(x')\geq f(x) - \phi(d_{\setDec}(x,x'))$.
    This includes the case of Hölder continuity, which has been used in a similar context before \citep{calliess2014conservative}.
\end{enumerate}
To keep the presentation focused and easy to follow, we do not use this additional flexibility in the present work, but everything that follows immediately applies to these more general cases.
\end{remark}
Our proposed modification applies to \emph{any} algorithm instantiating the generic SafeOpt strategy outlined in Section \ref{sec:problemSetting}.
For concreteness, we focus in the following on the original SafeOpt algorithm from \citet{sui2015safe}.
The resulting variant of LoSBO is described in detail in Algorithm \ref{alg:losbo}.
\begin{algorithm}[h!]
\caption{LoSBO}
\label{alg:losbo}
\begin{algorithmic}[1]
\Require Lipschitz constant $L$, algorithm to compute $\beta_t$, noise bound $E$, initial safe set $S_0$, safety threshold $h$
\State $Q_0(x) \gets \mathbb{R}$ for all $x \in D$ \Comment{Initialization of uncertainty sets}
\State $C_0(x) \gets [h,\infty)$ for all $x \in S_0$
\State $C_0(x) \gets \mathbb{R}$ for all $x \in D \setminus S_0$
\For{$t =1,2,\ldots$}
    \State $C_t(x) \gets C_{t-1}(x) \cap Q_{t-1}(x)$ for all $x \in D$ \Comment{Compute upper and lower bounds for current iteration}
    \State $\ell_t(x) \gets \min C_t(x)$, $u_t(x) \gets \max C_t(x)$ for all $x \in D$
    \If{$t > 1$} \Comment{Compute new safe set}
        \State $S_t = S_{t-1} \cup \{ x \in D \mid y_{t-1} - E - Ld(x_{t-1}, x) \geq h \}$
    \Else
        \State $S_1 = S_0$
    \EndIf
    \State $G_t \gets \{ x \in S_t \mid \exists x^\prime \in D \setminus S_t: \: u_t(x) - Ld(x,x^\prime) \geq h  \}$ \Comment{Compute set of potential expanders}
    \State $M_t = \{ x \in S_t \mid u_t(x) \geq \max_{x_S \in S_t} \ell_t(x_S) \}$ \Comment{Compute set of potential maximizers}
    \State $x_t \gets \argmax_{x \in G_t \cup M_t} w_t(x)$ \Comment{Determine next input}
    \State Query function with $x_t$, receive $y_t = f(x_t) + \epsilon_t$
    \State Update GP with new data point $(x_t,y_t)$, resulting in mean $\mu_t$ and $\sigma_t$
    \State Compute updated $\beta_t$
    \State $Q_t(x)=[\mu_t(x) - \beta_t \: \sigma_t(x), \mu_t(x) + \beta_t \: \sigma_t(x)]$ for all $x \in D$ 
\EndFor
\end{algorithmic}
\end{algorithm}

\begin{figure}[h!]
    \centering
    \includegraphics[width=0.8\textwidth]{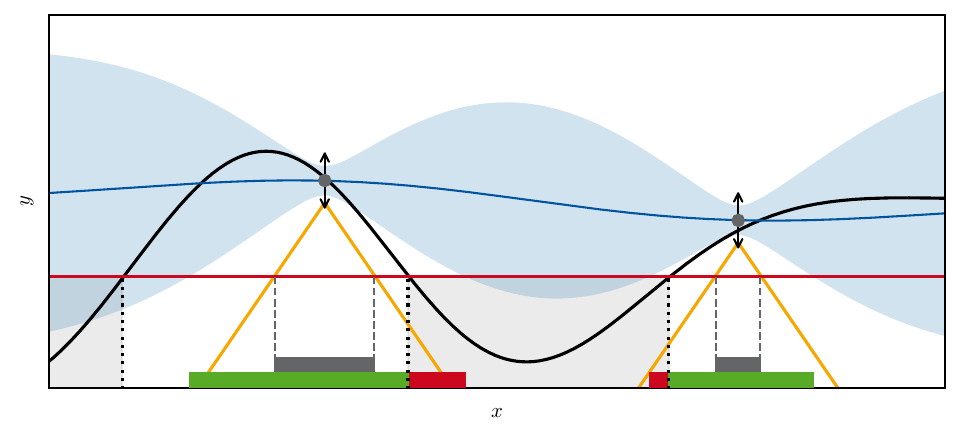}
    \caption{Illustration of LosBO being safe while a safety definition in based on leads to unsafe points in the safe set. The safe set of losbo (gray set) is determined by the constant $E$ (gray arrow) and the Lipschitz cone (orange). The GP mean and the confidence bounds are illustrated in blue. The points in the safe set given by the lower confidence bound are green if they are safe and red if they are unsafe.}
    \label{fig:losboIllustration}
\end{figure}

While LoSBO arises from a rather minor modification of the generic SafeOpt algorithm class (by changing the computation of the safe sets $S_t$), on a conceptual level significant differences arise.
Inspecting the proof of Proposition \ref{prop:losbo:safety} shows that the safety guarantee of LoSBO is \emph{independent} of the underlying model sequence $(\model_t)_t$.
As an important consequence, the choice of the uncertainty sets used in the optimization of the acquisition function cannot jeopardize safety.
One consequence is that the assumption that the target function $f$ is contained in the RKHS of the covariance function used in GP regression is not necessary anymore. 
In particular, \emph{in order to ensure safety, we need only Assumption \ref{assump:lipschitz}} together with a noise bound, \emph{and not Assumption \ref{assump:rkhsNormBound} anymore}.
Similarly, hyperparameter tuning is not an issue for safety, cf. also to our discussion in Section \ref{sec:uncertaintyBounds}.
Of course, an appropriate function model is important for good exploration performance, but this issue is now independent of the safety aspect.
As another, even more important consequence, in the context of the concrete LoSBO variant described in Algorithm \ref{alg:losbo}, the scaling parameter $\beta_t$ is now a proper tuning parameter. 
Modifying them, even online, in order to improve exploration does not infere with the safety requirements anymore.
This is in contrast to previous variants (and practical usage) of SafeOpt, where the scaling factors $\beta_t$ \emph{cannot} be freely tuning since they are central for the safety mechanism of these algorithms.
This aspect is illustrated in Figure \ref{fig:losboIllustration}.
In the situation depicted there, the uncertainty bounds do not hold uniformly, i.e., the target function is not completely covered by them, and deriving safety sets from these uncertainty bounds, regardless of whether to include the additional knowledge of the Lipschitz bound \citep{sui2015safe} or not \citep{berkenkamp2016bayesian}, results in potential safety violations.
However, since in LoSBO these bounds are ignored for the safe sets, this problem does not occur.

The original SafeOpt algorithm comes with conditional\footnote{By this we mean that the exploration guarantee given in \citet[Theorem~1]{sui2015safe} is conditional on the choice of the kernel. Inspecting the expression for the time $t^\ast$ in this latter result, one finds that this result requires appropriate growth behavior of the maximum information gain $\gamma_t$ in order to lead to non-vacuous exploration guarantees.} exploration guarantees.
Since our modification leading to LoSBO essentially separates the safety and exploration mechanisms, which makes the exploration guarantees from SafeOpt inapplicable in the present context.
An inspection of the proof of \citet[Theorem~1]{sui2015safe} shows that it cannot easily modified to apply to LoSBO again, since the argument used there relies on the GP model interacting with the safety mechanism\footnote{More precisely, with LoSBO one cannot ensure that the inequality in the last display in the proof of \citet[Lemma~7]{sui2015safe} holds.}.
Furthermore, we suspect that there exist pathological situations, where LoSBO fails to properly explore, however, we have not observed such a situation in our extensive experiments.
While providing (again conditional) exploration guarantees for LoSBO is an interesting question for future work, we argue that the present lack of such theoretical guarantees is not a problem for LoSBO and does not diminish the relevance and usefulness of this algorithm.
First, LoSBO shows excellent exploration performance, as demonstrated in the experiments described in the next section.
Second, since the scaling parameters $\beta_t$ (which have an important influence on the exploration performance) are proper tuning parameters in LoSBO, unsatisfying performance of the algorithm can be overcome by using this tuning knob.
We would like to stress again that in the previous variants of SafeOpt, one does not have this freedom since the scaling parameters need to lead to valid uncertainty sets.%

\begin{remark} \label{remark:noiseBoundLoSBO}
Proposition \ref{prop:losbo:safety} states that if $E\in\Rp$ is a bound on the noise magnitude, then LoSBO is safe.
Suppose we know that $|\epsilon_t|\leq B_\epsilon$ for some constant $B_\epsilon\in\Rp$, so one could set $E=B_\epsilon$, and assume that the bound $B_\epsilon$ is sharp.
For example, we might have $\epsilon_t \sim \frac12 \delta_{B_\epsilon} + \frac12 \delta_{-B_\epsilon}$, where $\delta_x$ is the Dirac distribution with atom on $x$.
If we choose $E=B_\epsilon$, then LoSBO is indeed safe according to Proposition \ref{prop:losbo:safety}, i.e., all inputs $x_t\in\setDec$ that are queried by the algorithm, we have $f(x_t)\geq \safeLevel$.
However, assume additionally that there are sizeable parts of the input space $\setDec$ where $f\approx \safeLevel$, and since the border of the safe sets will be in such a region, it is likely that inputs from this area will be queried.
This means that it is likely that measurements $y_t$ with $y_t < \safeLevel$ will be received - this happens if $f(x_t) \approx \safeLevel$ and $\epsilon_t \approx -B_\epsilon$.
While according to the formal model, such an input $x_t$ is safe, it looks to the user as if a safety violation occured.
Since in practice a detected (not necessarily real) safety violation might lead to some costly action (e.g., emergency stop), such a situation is undesirable.
In order to avoid, one can set $E=2B_\epsilon$. While this introduces some conservatism, it avoids the apparent safety violations just described - essentially, this option mitigates false alarms.
Whether to choose in the present situation $E=B_\epsilon$ or $E=2B_\epsilon$ (or even an option in between) is ultimately a practical question that needs to be addressed by the practioner using the algorithm.
\end{remark}

\subsection{Experimental evaluation} \label{sec:losboExperiments}
As discussed in Section \ref{sec:realBetaSafeOpt}, a meaningful comparison with SafeOpt implementations relying on heuristics is impossible, since the latter essentially address a different problem setting.
For this reason, we will compare LoSBO with Real-$\beta$-SafeOpt, which precisely adheres to the original SafeOpt setting.

\paragraph{Experimental setup}
For the empirical evaluations a frequentist approach will be used, since this is the most natural setting for SafeOpt-like algorithms, cf. \citet{Srinivas2010,fiedler2021aaai,fiedler2021cdc} for some discussion of this aspect.
This means that a target function is fixed, and the algorithms are run multiple times on this same function with independent noise realizations. 
In order to allow a clear evaluation of the performance of the algorithms, synthetic target functions will be used, and since we want to compare LoSBO with Real-$\beta$-SafeOpt - the latter requiring a target function from an RKHS and with a known RKHS norm upper bound - we generate target functions from an RKHS.
The frequentist setup is inherently worst-case, but for numerical experiments one has to restrict to finitely many RKHS functions.
Nevertheless, the RKHS functions used should be somewhat representative to give a meaningful indication of the algorithmic performance, in particular, any bias due to the function generating method should minimized.
In the following experiments, we sample functions from the pre RKHS, i.e., given a kernel $k$, we randomly choose some $M\in\Np$, $\alpha_1,\ldots,\alpha_M\in\R$ and $x_1,\ldots,x_M\in \setDec$ and then use $f = \sum_{i=1}^M \alpha_i k(\cdot, x_i) \in H_k$
as a target function, which works for \emph{any} kernel, cf. Section \ref{sec:rkhs}.
In the case of the squared exponential kernel, we also utilize the ONB described in \citet[Section~4.4]{SC08}, which we already used for some of the experiments in Sections \ref{sec:heuristicsProblem} and \ref{sec:problemWithRKHSnormBound}.
Generating RKHS functions with more than one method ensures more variety of the considered RKHS functions.
Moreover, with both approaches, the exact RKHS norm is available (and can be set by normalization), and the generated functions can be evaluated at arbitrary inputs.
Unless noted otherwise, we generate RKHS functions with an RKHS norm of $10$, i.e., we consider target functions $f\in H_k$ with $\|f\|_k=10$.
For a more thorough discussion of generating RKHS functions and subtle biases due to the chosen method, we refer to \citet{fiedler2021aaai}.

LoSBO and Real-$\beta$-SafeOpt work on arbitrary metric spaces, as long as a kernel can be defined on it.
Following the previous safe BO literature, we restrict ourselves to compact subsets of $\R^d$, and in this section we furthermore restrict us for simplicity to $d=1$.
In order to run LoSBO and Real-$\beta$-SafeOpt, we need a bound on the Lipschitz constant of the target function, as well as an initial safe set.
For the former, we restrict ourselves to kernels inducing continuously differentiable RKHS functions, since the latter are Lipschitz continuous due to the compact domain.
In order to determine a bound on the Lipschitz constant, we evaluate the target function on a fine discretization of the input domain, numerically compute an approximation of the Lipschitz constant, and multiply the result by 1.1 in order to counterbalance the discretization error.
Since the target functions are generated randomly, we compute an appropriate safety threshold for each function, so that some portions of the input space are safe, and some are unsafe.
This avoids trivial situations for safe BO.
More precisely, for a given target function $f$, we compute its empirical mean $\hat\mu(f)$ and empirical standard deviation $\hat{SD}(f)$ on a fine grid, and then set \expm{$\safeLevel=\hat\mu(f)-0.2\hat{SD}(f)$}.
Next, for each target function $f$ and safety threshold $\safeLevel$, we need to generate an initial safe set.
Similar to the choice of the safety threshold, trivial situations should be avoided, in particular, cases where no safe exploration is possible at all.
To achieve this goal, we first determine some $x_0 \in \argmax_{x\in \setDec} f(x)$,
then consider the set $\setDec \cap I_{x_0}$, where $I_{x_0}$ is the largest interval such that $x_0 \in I_{x_0}$ and $f\lvert_{I_{x_0}}\geq \safeLevel + E$, 
and finally one %
input is randomly selected from this set,
and the singleton set containing this latter input is then set as the initial safe set.
Using a singleton initial safe set is common in the literature on SafeOpt-type algorithms, cf. \citet{berkenkamp2016bayesian}.

The typical application scenario for SafeOpt-type algorithms is the optimization of some performance measure by interacting with a physical system.
In particular, each function query is relatively expensive, hence in these scenarios only a few function values are sampled.
Motivated by this, in all of the following experiments, for each target function, LoSBO and Real-$\beta$-SafeOpt are run for 20 iterations, starting from the same safe set.
For each target function, this is repeated \expm{10000 times} to allow a frequentist evaluation of the behavior.
Finally, each type of experiment is run with \expm{100} different randomly generated target functions.
To make runs with different target functions comparable, we evaluate the performance in a given run of a target function $f$ by
\begin{equation} \label{eq:experiments:evaluationMetric}
    \hat{f}^\ast_t = \frac{f\left(\argmax_{x\in S_t} \mu_t(x) \right)-\safeLevel}{f^\ast-\safeLevel},
\end{equation}
where $\mu_t(x)$ is the predictive mean (in LoSBO and Real-$\beta$-SafeOpt the posterior mean) at time $t\geq 1$, evaluated at input $x$,
and $f^\ast$ the maximum of $f$.
This metric will be averaged over all runs for a given $f$, and over all target functions, respectively, in the following plots.

\begin{figure}[t]
    \input{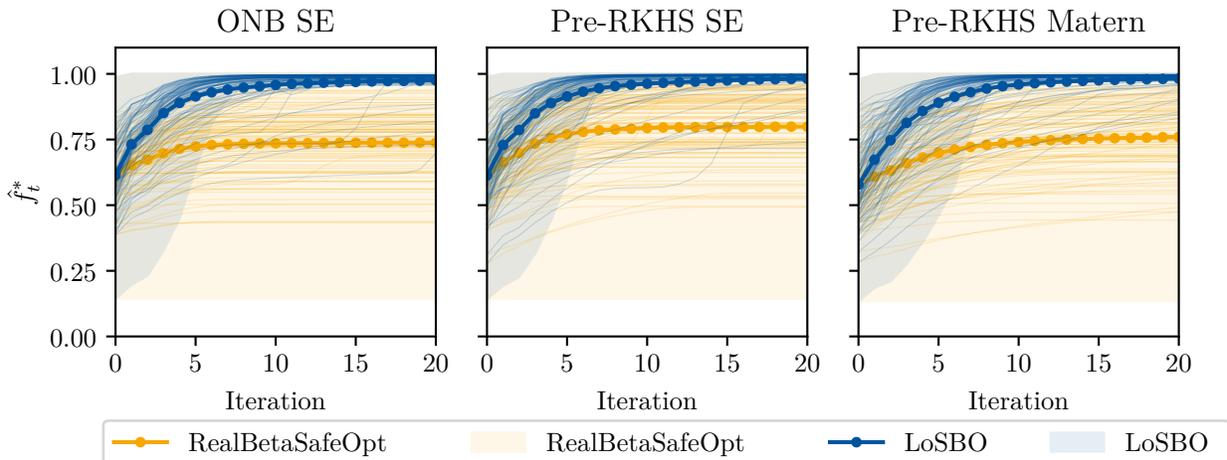}
    \caption{Comparison of LosBO and Real-$\beta$-SafeOpt in a well-specified setting. Thick solid lines are the means over all functions and repetitions, thin solid lines are the means over all repetitions for each individual function, shaded area corresponds to one standard deviation over all runs.}
    \label{fig:experiments:losboAndRealBetaSafeOptWellspecified}
\end{figure}

For simplicity, independent additive noise, uniformly sampled from $[-B_\epsilon,B_\epsilon]$, is used in all of the following experiments.
As is well-known, bounded random variables are subgaussian, and we can set \expm{$R=B_\epsilon$} in Real-$\beta$-SafeOpt.
Additionally, we choose \expm{$\delta=0.01$} and the true RKHS norm as the RKHS norm upper bound in Real-$\beta$-SafeOpt, unless noted otherwise.
Furthermore, we set the nominal noise variance equal to $R$ in both LoSBO and Real-$\beta$-SafeOpt.
Following the discussion in Remark \ref{remark:noiseBoundLoSBO}, we choose \expm{$E=2B_\epsilon$} in LoSBO.
Finally, a strategy to compute $\beta_t$ in LoSBO needs to be specified. 
Recall from Section \ref{sec:introducingLoSBO} that these scaling factors are now proper tuning parameters.
In all of the following experiments, we use \expm{$\beta \equiv 2$} in LoSBO, as this is a common choice in the literature on SafeOpt and GP-UCB.
Furthermore, choosing such a simple rule simplifies the experimental evaluation, since no additional tuning parameters or further algorithmic choices are introduced.
Finally, unless noted otherwise, in all of the following experiments \expm{$B_\epsilon=0.01$} is used.
For convenience, the experimental results are concisely summarized in Table \ref{tab:SafetyPerfomanceSafeOpt}.
\begin{table}[]
    \centering
    \begin{tabular}{lrrrrrr}\toprule
        Algorithm & SafeOpt $\beta=2$ & SafeOpt $B=2.5$ & SafeOpt $B=10$ & SafeOpt $B=20$ & LosBO \\
        \midrule
        Not started \% & 1.93 & 3.16 & 30.40 & 68.34 & 0.018 \\
        Safety violations \% & 3.95 & 0.859 & 0 & 0 & 0\\
        \makecell[l]{Safety violations \\ worst case \%} & 28.62 & 13.38 & 0 & 0 & 0\\
        Final performance \% & 88.75 & 88.76 & 82.45 & 76.69 & 90.90 \\
        \bottomrule
    \end{tabular}
    \caption{Safety-performance tradeoff in SafeOpt. We evaluated 100 Functions sampled from a SE-kernel with $B=10$. On each function we run 10000 times each algorithm starting from two initial safe points.}
    \label{tab:SafetyPerfomanceSafeOpt}
\end{table}
\paragraph{Well-specified setting}
We start by comparing LoSBO and Real-$\beta$-SafeOpt in a well-specified setting. This means that all the algorithmic parameters are set correctly, in particular, the covariance function used in GP regression is the kernel generating the RKHS from which the target functions are sampled.
In Figure \ref{fig:experiments:losboAndRealBetaSafeOptWellspecified}, the results for this setting are presented. 
The thick solid lines are the means over all \expm{100} functions and all of their repetitions, the fine lines are the means for each of the individual \expm{100} target functions (over \expm{10000} repetitions for each function), and the shaded area shows an interval of width one standard deviation around the mean, again over all functions.
In Figure \ref{fig:experiments:losboAndRealBetaSafeOptWellspecified}, top left, 
the results for functions from the Squared Exponential RKHS, sampled using the ONB approach, are displayed.
Interestingly, LoSBO exhibits superior performance compared to Real-$\beta$-SafeOpt, despite providing the latter algorithm with the correct ingredients (Lipschitz bound, kernel, RKHS norm bound, noise variance).
In Figure \ref{fig:experiments:losboAndRealBetaSafeOptWellspecified}, top right,
the results for functions sampled from the Squared Exponential pre RKHS are displayed.
While essentially no difference in performance is noticable for LoSBO, 
Real-$\beta$-SafeOpt appears to perform slightly better compared to target functions generated using the ONB approach.
A potential explanation lies in the shapes of functions that typically arise in the two different sampling methods.
As observed in \citet{fiedler2021aaai}, functions sampled from the ONB look more "bumpy" compared to pre RKHS functions, and appear to be more challenging for the uncertainty bounds.
Since Real-$\beta$-SafeOpt needs to exactly adhere to these bounds, its exploration performance is diminished.
In contrast, LoSBO behaves overly optimistic since $\beta_t \equiv 2$ is used, but the underlying RKHS functions have RKHS norm 10, cf. also the evaluations in \citet{fiedler2021aaai}.
It appears that this over-optimism leads to better performance, and since for safety LoSBO \emph{does not} rely on scaling factors $\beta_t$ that correspond to valid frequentist uncertainty sets, this over-optimism does not jeopardize safety.
Finally, in Figure \ref{fig:experiments:losboAndRealBetaSafeOptWellspecified}, lower left, the results for RKHS functions corresponding to a \expm{Matern-3/2} kernel and sampled with the pre RKHS approach are shown.
Qualitatively, we see the same picture, though the performance of both LoSBO and Real-$\beta$-SafeOpt appear to be slightly worse compared to the previous setting.
Intuitively, this is clear since Matern RKHS functions are generically less smooth than Squared Exponential RKHS functions, and both LoSBO and Real-$\beta$-SafeOpt rely on a Lipschitz bound, which in turn is related to regularity of functions.

\begin{figure}[t]
    \input{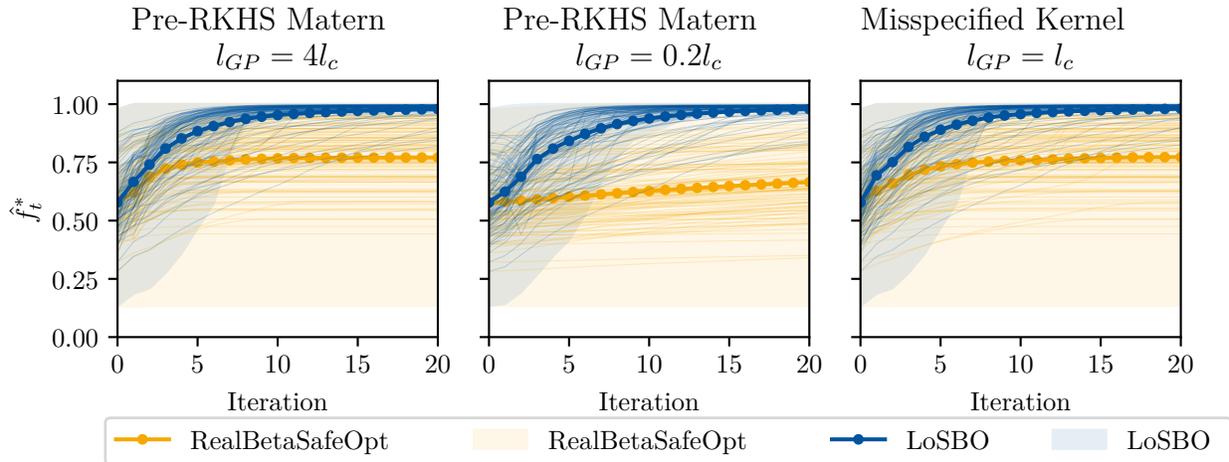}
    \caption{Comparison of LosBO and Real-$\beta$-SafeOpt in misspecified settings.}
    \label{fig:experiments:losboAndRealBetaSafeOptMisspecified}
\end{figure}

\paragraph{Misspecified setting}
We turn to misspecified settings, where the algorithmic parameters do not match the true setting of the target function.
This is particularly interesting in the present situation, since the underlying GP model does not impact the safety of LoSBO, and therefore becomes amenable to tuning.
The results of the experiments are displayed in Figure \ref{fig:experiments:losboAndRealBetaSafeOptMisspecified},
where as before the thick solid lines are the means over all 100 functions, 
the fine lines are the means for each of the individual 100 target functions, 
and the shaded areas show an interval of width one standard deviation around the mean, again over all functions.
We start with the length scale used in the kernel, which is arguably the most important type of hyperparameter in practice.
In Figure \ref{fig:experiments:losboAndRealBetaSafeOptMisspecified}, top left, we show the results for overestimating the length scale in GP regression, using Matern kernels.
More precisely, a Matern kernel is used both as the kernel for generating the target functions and as a covariance function in GP regression, but the length scale of the covariance function in GP regression is 4 times the ength scale used in the kernel to generate the target function.
The qualitative picture remains the same, though it appears that the performance of Real-$\beta$-SafeOpt suffers more than LoSBO from the misspecification.
More importantly, in this setting safety violations occur in \expm{12.57 \%} of all runs.
Moreover, for the worst-behaving target function \expm{943 out of 10000} runs lead to safety violations, which is unacceptable in a real-world use case.
In Figure \ref{fig:experiments:losboAndRealBetaSafeOptMisspecified}, top right, we show the complementary situation of underestimating the length scale in GP regression, again using Matern kernels.
The length scale of the covariance function used in GP regression is 0.2 times the length scale of the kernel that is used to generate the target function.
Again, the qualitative picture remains the same, but the performance degradation is worse for both algorithms in this case.
Finally, consider the case where a different kernel is used to generate the target functions than the covariance function in the GP regression.
We use a \expm{Matern-3/2} kernel to generate the target functions, and a \expm{Squared Exponential kernel} as covariance function in the GP regression, with \expm{the same length scale for both}.
The results are displayed in Figure \ref{fig:experiments:losboAndRealBetaSafeOptMisspecified}, lower left.
Interestingly, essentially no qualitative difference compared to the well-specified Matern case (Figure \ref{fig:experiments:losboAndRealBetaSafeOptWellspecified}, lower left) can be noticed.
We suspect that this is due to the correct specification of the length scale, which in the present setting is more important than the kernel misspecification.
\section{LoS-GP-UCB} \label{sec:highdim}
The central computational step in SafeOpt-type algorithms, cf. Algorithm \ref{alg:safeopt}, of which LoSBO (Algorithm \ref{alg:losbo}) is one variant, 
is the optimization of the acquisition function over the expander and maximizer sets.
Inspecting the definition of the latter two sets, cf. Section \ref{sec:problemSetting}, makes it clear that computing these sets requires a discrete input set $\setDec$.
Since for many typical application scenarios at least parts of the input set will be continuous (e.g., if the optimization variables include a physical parameter that can vary continuously), this means that some form of discretization becomes necessary before a SafeOpt-type algorithm is applicable.
Typically, one uses equidistant gridding of the (continuous parts of the) input set as a discretization, cf. \citet{berkenkamp2016bayesian} for a typical example.
As a result, SafeOpt-type algorithms become impractical for even moderate dimensions, e.g., $\setDec\subseteq\R^d$ for $d>3$ \citep{kirschner2019adaptive,sukhija2023gosafeopt}, and as a member of this class, LoSBO inherits this limitation.
In this section, we present and investigate an approach to overcome this limitation, addressing our objective (O3).
Instead of adapting existing solution approaches like \citet{duivenvoorden2017constrained,kirschner2019adaptive,sukhija2023gosafeopt}, we suggest a pragmatic and straightforward variant that is motivated by three observations.

\begin{figure}[t]
    \includegraphics[width=0.5\textwidth]{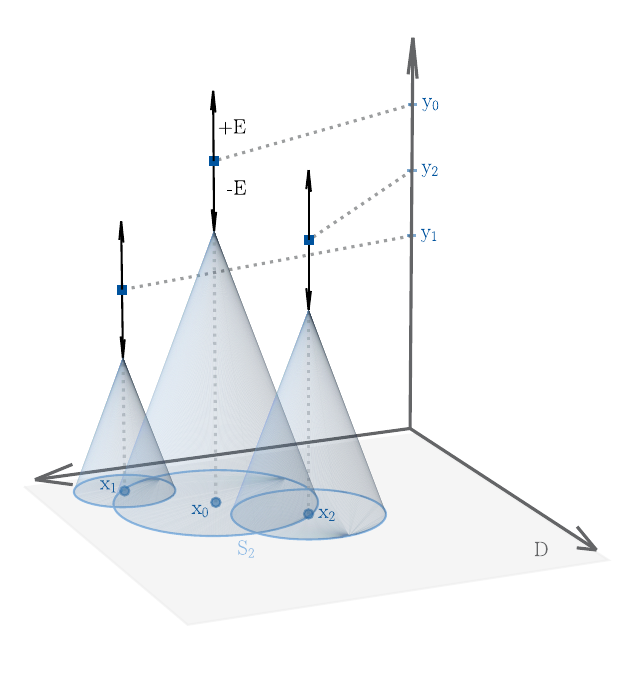}
    \includegraphics[width=0.5\textwidth]{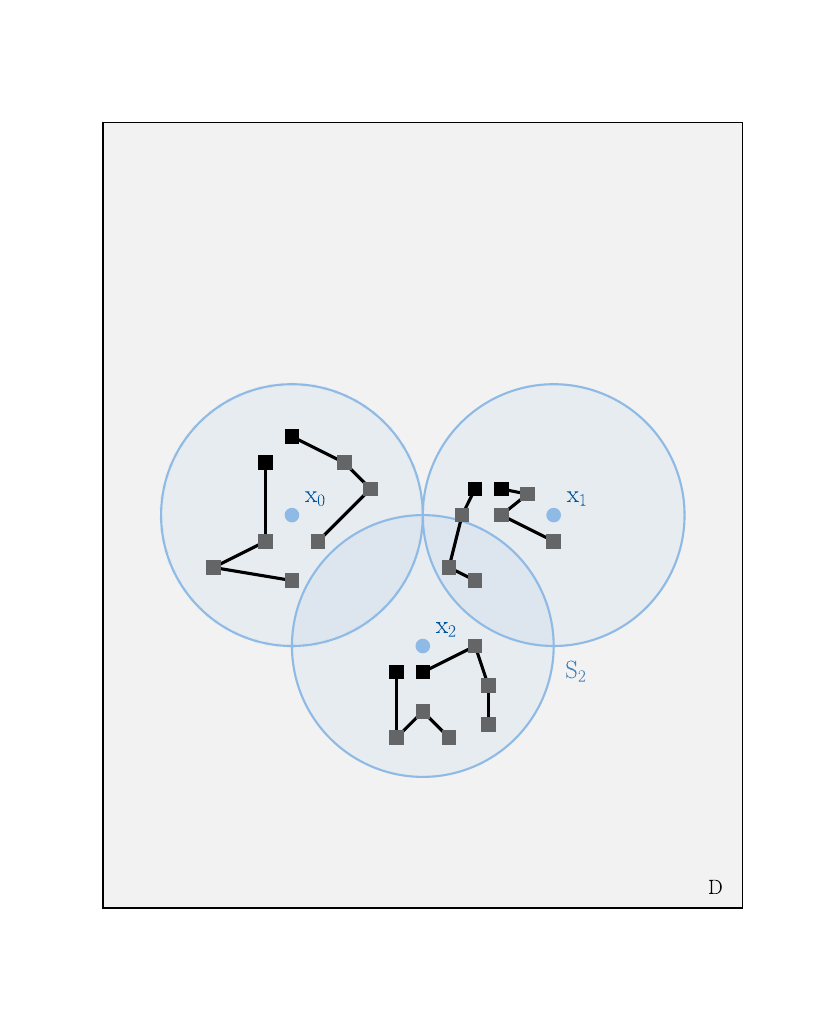}
    \caption{Illustration of safe sets for a 2-dimensional input set. Left: Safe set in LoSBO resulting from three function evaluations.
    Right: Illustration of the acquisition function optimization in LoS-GP-UCB over a safe set resulting from four function evaluations. Two initial guesses for the local optimization are used for each of the balls that span the safe set according to \eqref{eq:safeSetDecomp}.}
    \label{fig:illustrationLoS-GP-UCB}
\end{figure}

\emph{First}, safety in SafeOpt-type algorithms is ensured by restricting the optimization of the acquisition function to (subsets of) safe sets, i.e., sets $S\subseteq \setDec$ such that $f\lvert_S \geq \safeLevel$, where $f$ is the unknown target function.
In other words, as long as we ensure that the acquisition function optimization is restricted to such sets $S$, the resulting algorithm will be safe, no matter how this optimization is performed or whether an additional restriction is added (as in SafeOpt, where the optimization is only over expander and maximizer sets).
In the case of LoSBO, these safe sets are of a particularly simple form, since they are the union of closed balls in a metric space, cf. Figure \ref{fig:illustrationLoS-GP-UCB}, left, for an illustration for the case $\setDec\subseteq\R^2$.
Since in typical application scenarios of SafeOpt-type algorithms, the number of input queries is relatively low, and hence the aforementioned union is only over relatively few sets.

\emph{Second}, a discrete input set for SafeOpt-type algorithms is necessary due to the involved definition of the expander and maximizer sets, which in turn are defined to guarantee proper exploration in the original SafeOpt setting \citep{sui2015safe}.
However, for concrete practical problems, such an underexploration might not pose a severe challenge, and one can simply make an existing BO algorithm safe by restricting the optimization of the acquisition function to a safe set $S$ as described above.
In fact, the original SafeOpt paper \citet{sui2015safe} already discussed a safe variant of GP-UCB \citep{Srinivas2010}.
This indicates that it might be possible to avoid the complicated sets involved in SafeOpt-type algorithms, and still attain practically useful exploration performance.

\emph{Third}, in the current practice of BO, moderately high dimensions of the input space are not a problem for modern BO algorithms\footnote{Here we are referring to handling the dimensionality within the algorithm, in particular, optimization of the acquisition function, and not the exploration performance. Of course, the latter poses challenges, especially if not enough structural or qualitative prior knowledge is encoded in the BO function model.}
Typically, acquisition functions are optimized by running a local optimization method (usually gradient-based) from several initial guesses (which might be random, or based on heuristics).
In particular, no gridding is necessary, and this strategy can deal even with moderate high dimensions since local search methods behaves well despite increasing dimensionality.
In fact, state-of-the-art BO libraries like BOTorch \citep{balandat2020botorch} implement exactly this approach as the default option.

Based on these three observations, we now propose a straightforward safe BO algorithm that works even in moderate input dimensions, is compatible with modern BO libraries, and retains all the safety guarantees from LoSBO.
In Section \ref{sec:los-gp-ucb:alg} we describe this algorithm in detail and provide some discussion.
In Section \ref{sec:los-gp-ucb:experiments} we evaluate the algorithm empirically and compare it to LoSBO.

\subsection{Algorithm} \label{sec:los-gp-ucb:alg}
We now introduce a practical safe BO algorithm that retains the favorable properties of LoSBO, but also works in moderately high dimensions.
Consider the setting of LoSBO as described in Section \ref{sec:introducingLoSBO}.

Motivated by the preceding discussion, we start with a standard GP-based BO algorithm that does not need expander and maximizer sets (or similar complicated sets that require discretization).
Due to its relation to SafeOpt, we choose GP-UCB \citep{Srinivas2010} for this task, so at step $t\geq 1$, the next input is
\begin{equation}
    x_{t+1} = \argmax_{x \in \setDec} \mu_t(x) + \beta_t \sigma_t(x),
\end{equation}
for an appropriate scaling factor $\beta_t\in\Rp$.
As usual, ties are broken arbitrarily.
Using the (scaled) posterior variance as the acquisition function would be even closer to SafeOpt, but numerical experiments indicate that GP-UCB performs slightly better in this context.
Next, we restrict the acquisition function optimization to the safe sets $S_t$ as defined for LoSBO in \eqref{eq:losbo:safeset},
\begin{equation}
     x_{t+1} = \argmax_{x \in S_t} \mu_t(x) + \beta_t \sigma_t(x).
\end{equation}
Observe now that 
\begin{equation} \label{eq:safeSetDecomp}
    S_t = \bigcup_{j=1}^{N_t} \bar{B}_{r_j}(z_j)
\end{equation}
for some $N_t\in\Np$, $r_1,\ldots,r_{N_t}\in\Rp$ and $z_1,\ldots,z_{N_t}\in\setDec$,
and $\bar{B}_r(z)=\{ x \in \setDec \mid d_\setDec(z,x)\leq r\}$ is the closed ball with radius $r\in\Rp$ and center $z\in\setDec$ in the metric space $\setDec$.
For example, if the initial safe set has only one element $x_0$ and no input is repeatedly sampled, then $N_t=t+1$, $z_1=x_0$ and $z_j=x_{j-1}$ for $j=2,\ldots,t+1$.
Using the decomposition \eqref{eq:safeSetDecomp}, we now have
\begin{equation}
     x_{t+1} = \argmax_{j=1,\ldots,N_t} \argmax_{x \in \bar{B}_{r_j}(z_j)} \mu_t(x) + \beta_t \sigma_t(x).
\end{equation}
Each of the inner optimization problems $\max_{x \in \bar{B}_{r_j}(z_j)} \mu_t(x) + \beta_t \sigma_t(x)$, $j=1,\ldots,N_t$, is a maximization problem over the convex sets $\bar{B}_{r_j}(z_j)$, and each of these inner problems are independent.
In particular, these optimizations can be trivially parallelized.
In practice, one usually has $D\subseteq \R^d$ (often with a simple geometry) and a differentiable covariance function $k$, so it is possible to use a gradient-based local optimization method started from multiple initial guesses.
This is illustrated in Figure \ref{fig:illustrationLoS-GP-UCB}, right.
All of these multistarts are independent, and can therefore be also parallelized.
Altogether, we arrive at Algorithm \ref{alg:los-gp-ucb}, which we call Lipschitz-only Safe Gaussian Process Upper Confidence Bound (LoS-GP-UCB) algorithm in the following.
\begin{algorithm}[t]
\caption{LoS-GP-UCB}
\label{alg:los-gp-ucb}
\begin{algorithmic}[1]
\Require Lipschitz constant $L$, algorithm to compute $\beta_t$, noise bound $E$, initial safe set $S_0$, safety threshold $h$
\State Compute $N_0$, $z_1,\ldots,z_{N_0}$, $r_1,\ldots,r_{N_0}$, $\beta_0$
\For{$t =1,2,\ldots$}
    \State $ x_{t} = \argmax_{j=1,\ldots,N_{t-1}} \argmax_{x \in \bar{B}_{r_j}(z_j)} \mu_{t-1}(x) + \beta_{t-1} \sigma_{t-1}(x).$ 
        \Comment{Determine next input}
    \State Query function with $x_t$, receive $y_t = f(x_t) + \epsilon_t$
    \State Update GP with new data point $(x_t,y_t)$, resulting in mean $\mu_t$ and $\sigma_t$
    \State Compute updated $\beta_t$
    \State Compute $N_t$ and add new $z_j$, $r_j$
\EndFor
\end{algorithmic}
\end{algorithm}
LoS-GP-UCB can be easily implemented with state-of-the-art BO libraries. 
For the numerical experiments described in the next section, we have chosen BoTorch \citep{balandat2020botorch}, which allows an easy parallel implementation of the acquisition function optimization.

Finally, LoS-GP-UCB retains the safety guarantees from LoSBO.
In particular, the scaling factors $(\beta_t)_t$ remain tuning factors, and the safety of LoS-GP-UCB is independent of their choice.
Furthermore, safety of LoS-GP-UCB \emph{does not require Assumption \ref{assump:rkhsNormBound}}.
Similarly, Remarks \ref{remark:generalizations} and \ref{remark:noiseBoundLoSBO} also apply to LoS-GP-UCB.

\subsection{Experimental evaluation} \label{sec:los-gp-ucb:experiments}
Completely analogous to the case of Real-$\beta$-SafeOpt and LoSBO, a frequentist setup will be used, i.e., the algorithm will be run on a fixed target function for many independent noise realizations.
In the experiments two aspects will be investigated.
First, LoS-GP-UCB will be compared with Real-$\beta$-SafeOpt and LoSBO.
Second, we apply LoS-GP-UCB to several benchmark functions with moderate input dimensions.

\begin{figure}[h!]
    \centering
    \input{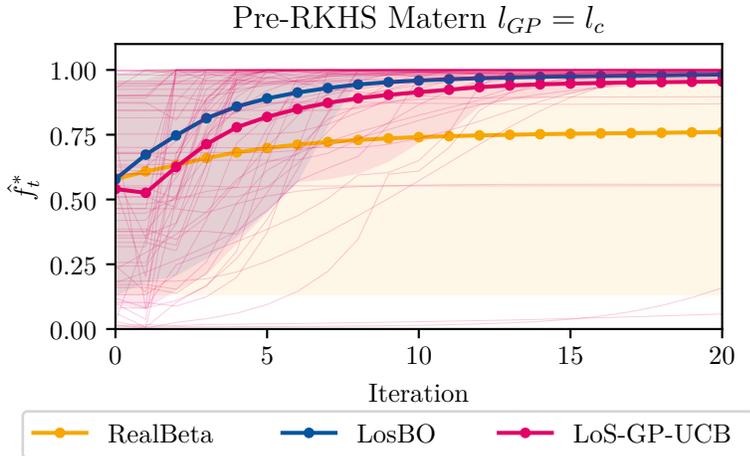}
    \caption{Comparison of LoS-GP-UCB to Real-$\beta$-SafeOpt and LoSBO.}
    \label{fig:losgpucbRKHS}
\end{figure}

We start with the comparison to Real-$\beta$-SafeOpt and LoSBO.
Since these two algorithms rely on a discrete input space, this comparison necessarily has to be performed on functions with a low dimensional input. 
We choose essentially the same experimental settings as in Section \ref{sec:los-gp-ucb:experiments} and consider only the well-specified case.
In particular, as for LoSBO, we use $\beta_t \equiv 2$ in LoS-GP-UCB, and we consider one-dimensional RKHS functions.
The algorithms are evaluated on \expm{100} target functions sampled from the pre RKHS corresponding to a \expm{Matern-3/2} kernel, and for each function, we run the algorithms \expm{1000} times\footnote{The qualitative picture is already clear for 1000 repetitions, hence we choose to save computational resources and do not use 10000 repetitions as in Section \ref{sec:los-gp-ucb:experiments}.}. 
The results of this experiment are shown in \ref{fig:losgpucbRKHS}, where we again use the evaluation metric \eqref{eq:experiments:evaluationMetric}.
Thick solid lines are the means over all functions and repetitions, the shaded areas are correspond to one standard deviation from the mean (again over all functions and realizations).
To avoid clutter, the means for each individual function are plotted only for LoS-GP-UCB (thin lines).
It is clear from Figure \ref{fig:losgpucbRKHS} that Los-GP-UCB performs only slightly worse than the original LosBO algorithm, but still outperforms Real-$\beta$-SafeOpt. 
This outcome indicates that LoS-GP-UCB is not severely affected by the under-exploration problem described for a safe variant of GP-UCB in \citet{sui2015safe}.
We suspect that similar to LoSBO, this is due to the overoptimism resulting from setting $\beta_t\equiv 2$, which corresponds to moderately aggressive exploration.

Let us turn to the evaluation of LoS-GP-UCB on several benchmark functions with moderate to high input dimensions.
As test functions, we use the standard benchmarks Camelback (2d) and Hartmann (6d). 
Furthermore, similar to \cite{kirschner2019adaptive}, we use in addition a Gaussian function $f(x)= \exp{(-4 \lVert x \rVert_2^2)}$ in ten dimensions (10d) as a benchmark. 
For the Camelback and Hartmann functions, we choose a random initial point in the safe set. For the Gaussian function, we choose a random safe set from the level $f(x_0) = 0.4$. 
We assume uniformly bounded noise with a noise level of \expm{$0.01$}. 
In this synthetic setting the Lipschitz constant $L$ is determined by evaluating the function on a fine grid. As a model we use a \expm{Squared Exponential kernel} with output variance set to \expm{1} and length scale set to \expm{$1/L$}. 
For the prior mean we choose \expm{0.5} as the function values are between \expm{0} and \expm{1}. 
Finally, in all these settings, we compare LoS-GP-UCB to random search, and run both algorithms from the same initial safe set for \expm{100} iterations, repeating this \expm{100} times (for different random choices of the initial safe set).

\begin{figure}
    \includegraphics[width=\textwidth]{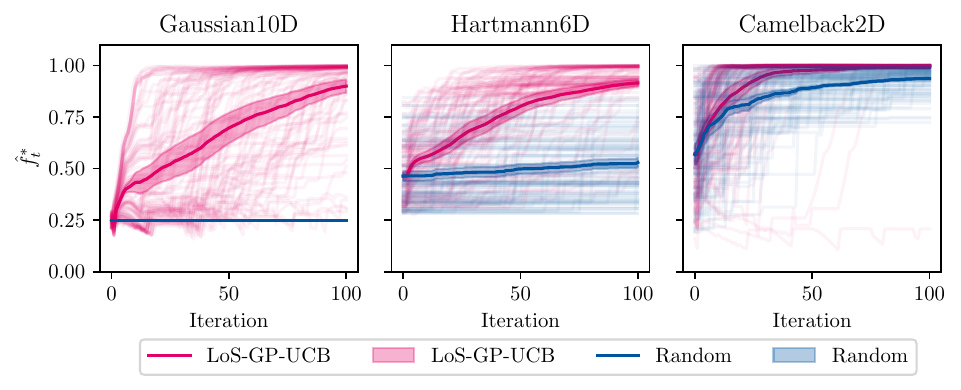}
    \caption{Evaluating LoS-GP-UCB (\expm{violet}) on three benchmark functions and comparison to random search (\expm{yellow}).
    Thick line are the means over all runs, thin lines are individual runs.
    }
    \label{fig:losucb}
\end{figure}

\section{Conclusion} \label{sec:conclusion}
In this work, we are concerned with practically relevant safety aspects of the important class of SafeOpt-type algorithms.
We identified the use of heuristics to derive uncertainty bounds as a potential source of safety violations in the practical application for these algorithms.
This prompted us to use modern, rigorous uncertainty bounds in SafeOpt, which allowed us to numerically investigate the safety behavior of this algorithm.
Furthermore, we identified the knowledge of an upper bound on the RKHS norm of the target function as a serious obstacle to reliable real-world applicability of SafeOpt-type algorithms.
In turn, we proposed LoSBO, a BO algorithm class relying only on a Lipschitz bound and noise bound to guarantee safety.
Numerical experiments showed that this algorithm is not only safe, but also exhibits superior performance.
Ongoing work is concerned with implementing the presented algorithms for safe learning in an automotive context, as well as providing exploration guarantees for LoSBO.
Furthermore, we expect that the approach outlined in Section \ref{sec:losbo} applies to most SafeOpt variants. Therefore, the derivation, implementation, and evaluation of the corresponding LoSBO-type algorithms for these variants is another interesting direction for future work.
Since we expect that the approach outlined in Section \ref{sec:losbo} applies to most SafeOpt variants, the development of corresponding LoSBO-type algorithms for these variants is another interesting direction for future work.
Finally, our findings in combination with evidence in the literature that SafeOpt and related algorithms have been successfully used in various applications indicate that this algorithm class does not ensure hard safety constraints (in practice), but instead yields ``cautious'' behavior. The precise connection to conservative bandits and existing cautious BO approaches is another interesting topic for further investigations.

\subsubsection*{Broader Impact Statement}
This work is concerned with safety issues of a popular BO algorithm that has already found numerous applications in real-world scenarios.
Henceforth we contribute to the improved safety and reliability of machine learning methods for real-world applications.
Furthermore, we expect no adverse societal impact of our work.

\subsubsection*{Acknowledgments}
We thank Paul Brunzema and Alexander von Rohr for very helpful discussions and Sami Azirar for support when generating plots. This work was performed in part within the Helmholtz School for Data Science in Life, Earth and Energy (HDS-LEE). Furthermore, the research was in part funded by the German Federal Ministry for Economic Affairs and Climate Action (BMWK) through the project EEMotion. Computations were performed with computing resources granted by RWTH Aachen University under project rwth1459.

\bibliography{main}
\bibliographystyle{tmlr}

\end{document}